\documentclass[lettersize,journal]{IEEEtran}
\usepackage{amsmath,amsfonts}
\usepackage{algorithm}
\usepackage{algpseudocode} 
\usepackage{array}
\usepackage[caption=false,font=normalsize,labelfont=sf,textfont=sf]{subfig}
\usepackage{textcomp}
\usepackage{stfloats}
\usepackage{url}
\usepackage{hyperref}
\usepackage{verbatim}
\usepackage{graphicx}
\usepackage{cite}
\usepackage{microtype}
\usepackage{booktabs}
\usepackage{arydshln}
\usepackage{multirow}
\usepackage{xcolor} 
\definecolor{red}{RGB}{166, 25, 46}  
\definecolor{blue}{RGB}{2, 77, 127} 
\definecolor{green}{RGB}{43, 127, 62}
\definecolor{purple}{RGB}{178,48,167} 
\begin{document}

\title{LLaVA-Assessor: Building the Foundation LMM For Visual Quality Assessment}

\author{
            Ziheng Jia, Zicheng Zhang, Jiaying Qian, Guangtao Zhai,~\IEEEmembership{Fellow, IEEE}, 
            Xiongkuo Min,~\IEEEmembership{Member, IEEE}
\thanks{This paper is an extension of our two previous conference papers, which were published in the Proceedings of the 33rd ACM International Conference on Multimedia~(ACM MM 2025)\textbf{[DOI: 10.1145/3746027.3754696]} and the Proceedings of the 40th AAAI Conference on Artificial Intelligence~(AAAI 2026, Oral)\textbf{[DOI: 10.1609/aaai.v40i27.39386]}, respectively.}
\thanks{
Ziheng Jia, Jiaying Qian, Xiongkuo Min and Guangtao Zhai are with the Institute of Image Communication and Information Processing, Shanghai Jiao Tong University, Shanghai 200240, China~(e-mail:\{jzhws1, 2022qjy, zhaiguangtao, minxiongkuo\}@sjtu.edu.cn)
}
\thanks{
Zicheng Zhang is with the Shanghai Artificial Intelligence Laboratory, Shanghai 200032, China~(e-mail: zhangzicheng@pjlab.org.cn)
}
}




\maketitle

\begin{abstract}
Aligning with the human visual system~(HVS) in perceiving and evaluating the quality of visual signals is a central objective of machine-vision-based visual quality assessment systems. With the rapid progress of large multi-modal models~(LMMs), visual question answering provides a promising paradigm for building unified foundation models for visual quality assessment under multi-modal and multi-task scenarios. Inspired by the classical ``perception-decision" process in HVS-based quality evaluation, we formulate visual quality assessment for LMM-based machine vision as two complementary tasks: ``quality interpretation'' and ``quality scoring". Centered on these objectives, we propose \textbf{\textit{LLaVA-Assessor}}, a unified
data construction and model training system. To support multi-modal inputs, we design an adaptive model architecture that enables efficient processing of both images and videos. For data construction, we develop rigorous human annotation protocols and a novel machine-synthesis-dominated data expansion pipeline to build a large-scale and high-quality datasets. Furthermore, we introduce a simple yet effective prompt disentanglement strategy to alleviate training-objective confusion in multi-task learning, thereby enabling stable and coherent joint training. The resulting all-in-one LMM \textbf{\textit{LLaVA-Assessor-GIGA}} achieves superior performance on $11$ image/video quality scoring test sets and $4$ visual quality interpretation benchmarks. Extensive results demonstrate the effectiveness of integrating structured data construction, adaptive model design, and multi-task joint training for automated visual quality assessment. Our work provides compelling insights for developing foundation LMMs for automatic visual quality assessment. Project page at \url{https://github.com/jzhws/LLaVA-Assessor}.

\end{abstract}


\section{Introduction}
With the rapid advancement of large multi-modal models~(LMMs)~\cite{achiam2023gpt,gao2023llama,zhang2024video,liu2024visual,bai2025qwen2}, computer vision has undergone a substantial paradigm shift. LMM-based visual question answering~\cite{antol2015vqa} has emerged as an effective interface for cross-modal understanding and alignment, and has been widely adopted in diverse applications. In particular, visual-language instruction tuning~\cite{liu2024visual}, supported by large-scale multi-modal instruction databases~(MIDBs), has significantly enhanced the capability of LMMs in complex semantic reasoning and interactive visual understanding. In parallel, extending visual question answering to low-level vision tasks has attracted increasing attention, offering a promising route toward unifying perceptual analysis and language-based reasoning. Visual quality assessment is a representative task in this direction, where the model is expected to not only perceive low-level visual degradations but also interpret and quantify their perceptual impact.
\begin{figure}
    \centering
    \includegraphics[width=\linewidth]{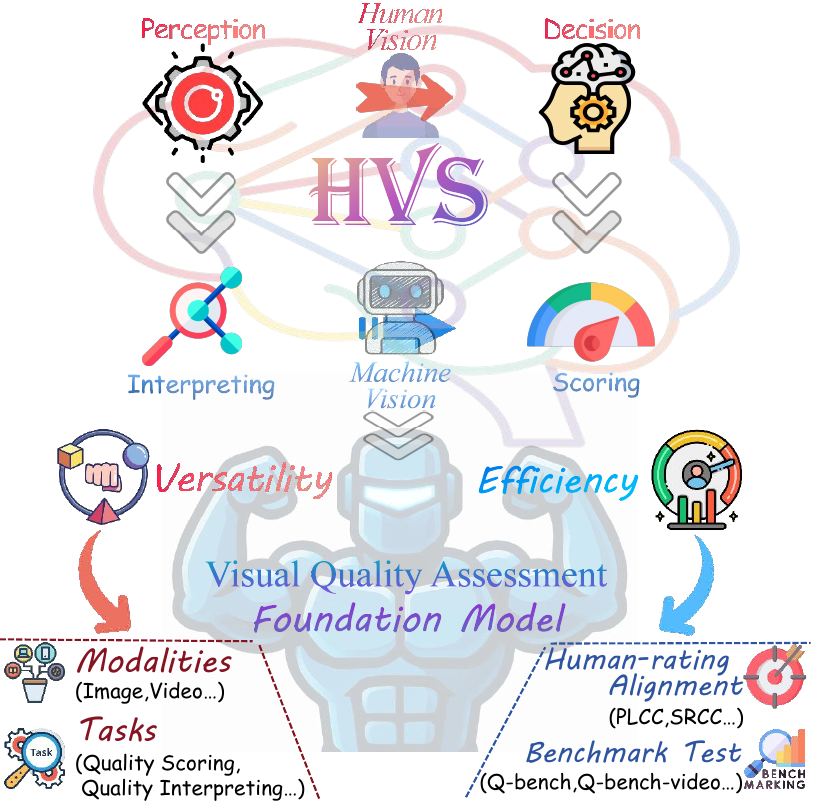}
    \vspace{-12pt}
    \caption{The motivation guideline of our work. The ``\textbf{perception-decision}" mechanism in the HVS system outlines the key tasks for machine-vision-based models, namely ``quality interpreting" and ``quality scoring". Building upon this, a visual quality assessment foundation LMM should focus on the model's \textbf{versatility} and \textbf{efficiency}. A unified model, when deployed once, should be capable of handling multi-modal, multi-task scenarios, achieving performance that aligns with human perception and is comparable to or surpasses that of proprietary models on benchmark tests for the corresponding tasks.}
    \vspace{-12pt}
    \label{fig:spotlight}
\end{figure}

A comprehensive assessment of a visual signal, such as an image or a video, should be consistent with the general processing workflow of the human visual system~(HVS), namely the ``\textbf{perception–decision}" paradigm~\cite{mazurek2003role}. In the perception stage, the HVS forms a holistic impression of the signal and derives an intuitive global quality judgment. In the subsequent decision stage, higher-level neural systems \textbf{rationalize} this perceptual judgment, for example, by assigning a Mean Opinion Score~(MOS) to the visual signal. Formulating this paradigm for machine-vision-based quality assessment naturally leads to two complementary tasks: \textbf{(1) the quality interpretation task, where the machine produces fine-grained qualitative descriptions of the visual input, analogous to human perception; (2) the quantitative quality scoring task, wherein the machine aims to predict quantitative visual quality scores that closely align with the decision stage.} Accordingly, we define the objective of developing a visual quality assessment foundation LMM~(the guideline is illustrated in Fig.~\ref{fig:spotlight}) as \textbf{comprehensively mimicking the HVS ``perception–decision" mechanism}. The desired model should exhibit both ``\textbf{versatility}" and ``\textbf{efficiency}". Specifically, once deployed, the unified model can handle multi-modal and multi-task inputs~(versatility) while achieving performance comparable to models tailored to individual tasks~(efficiency).


Achieving the above targets requires several essential considerations. First, image and video quality differ in emphasis: while video quality involves spatial visual quality, it further depends on temporal quality factors, such as stuttering and temporal discontinuities, thereby imposing stronger requirements on \textbf{frame processing}. Second, the foundation LMM should be closely \textbf{aligned with human perception}; hence, expert-guided annotation of high-quality training MIDB is fundamental to establishing such alignment. Third, given the substantial human annotation cost, it is necessary to scale up the MIDB with data primarily \textbf{synthesized by machines} to satisfy large-scale training demands. Finally, effective training requires the principled integration of multi-task and multi-modal data into a coherent \textbf{joint training}. We detail these considerations and the corresponding solutions below.


\noindent \textbf{\textit{The model structure.}} 
To accommodate the distinct inputs of images and videos, the base model must possess an adaptable architecture. For images that primarily focus on the pixel-level spatial feature, the input should preserve the original spatial resolution. Additionally, video quality assessment emphasizes temporal processing among densely sampled frames. Therefore, it is essential to incorporate a module that can capture temporal dynamics. Furthermore, considering the convenience of mixed training, the model architecture must support gated selection based on the input modality.

\begin{figure}
    \centering
    \includegraphics[width=\linewidth]{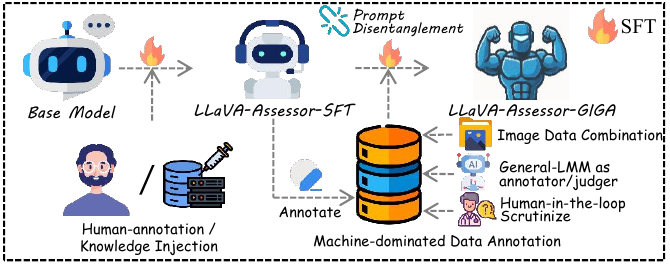}
    \vspace{-20pt}
    \caption{An overview of the technical map. We begin by conducting rigorous human annotation and training the intermediate \textbf{LLaVA-Assessor-SFT} using the human-annotated data for subsequent machine-dominated data expansion. Next, we employ techniques such as \textbf{general-LMM-as-annotator/judger} and \textbf{human-in-the-loop scrutinizing}, while also \textbf{integrating} high-quality \textbf{image quality assessment training data}. By utilizing the \textbf{prompt disentanglement} strategy, we achieve collective training, resulting in the \textbf{LLaVA-Assessor-GIGA}, a powerful unified visual quality assessment foundation LMM. }
    \label{fig:HVS}
    \vspace{-0.5cm}
\end{figure}
\noindent \textbf{\textit{Human labeled data construction.}} 
High-quality human-annotated data constitute the cornerstone for aligning the foundation LMM with human perception. For the quality scoring task, large-scale subjective ratings collected from human experts are indispensable. The resulting MOS values directly reflect perceived visual quality and serve as the \textbf{golden labels} for subsequent machine learning and evaluation.
For the quality interpretation task, human annotations are expected to complement MOS values by providing perceptual information that scalar scores alone cannot capture. Specifically, such annotations should cover:~(1) \textbf{quality factor interpretation}, which provides qualitative descriptions of specific quality attributes, including overall rendering quality as well as spatial characteristics and, for videos, temporal characteristics such as clarity, color, brightness, flicker, and stuttering; and~(2) \textbf{spatiotemporal fine-grained descriptions}, which characterize in-context visual signal properties, including spatial distortion regions, temporal degradation intervals, and spatiotemporal dynamics. These fine-grained perceptual cues are often difficult for general LMMs to perceive and model accurately.



\noindent \textbf{\textit{Scaling-up the training MIDB.}} 
General foundation LMMs typically require fine-tuning on MIDBs containing millions of samples to achieve comprehensive  functionality. Similarly, in the domain of visual quality assessment, the scaling-up of synthesis training data is critically important. This naturally raises a significant concern: \textbf{how can the quality of machine-driven annotations be ensured?}
Firstly, current high-performing general LMMs~(such as \textit{Gemini} or \textit{GPT-4o}) have limited accuracy in perceiving low-level  visual distortions. Directly using general LMMs to annotate low-level visual quality may often result in considerable bias. Therefore, we argue that the machine annotators involved in data synthesis should be domain-specific models. The expert model generates multiple sampled responses for the same query, which are then filtered through rejection sampling by an LMM-as-judger framework. Samples that remain difficult to judge by general LMMs are subjected to additional human-in-the-loop scrutiny. Moreover, considering that general LMMs have already demonstrated maturity in addressing higher-level tasks, it is also feasible to leverage these strengths by directly employing these models for relatively high-level annotation.

\noindent \textbf{\textit{Towards effective joint-training.}} 
A significant challenge also lies in the \textbf{target confusion}~(elaborated in Sec.~\ref{setting}) that arises in joint training. To address this issue, we propose an efficient strategy termed \textbf{prompt disentanglement}, whereby discarding the specialized semantic guiding prompt for the quality scoring task; instead, the model directly predicts the quality score based solely on the visual tokens. This approach effectively mitigates performance degradation caused by training-target confusion during joint training.
In summary, our core contributions are as follows and depicted in Fig.~\ref{fig:HVS}:

\begin{itemize} 
\item We propose a complete, high-quality MIDB construction pipeline for low-level visual quality assessment. This pipeline begins with human experts' annotation, thereby obtaining a large-scale MIDB aligned with human perception. Subsequently, we obtain the \textbf{\textit{LLava-assessor-SFT}} upon training on the human-annotated data, which serves as the expert model for machine annotation. we further implement the efficient dataset expansion by employing \textbf{rejection sampling fine-tuning~(RFT)}, along with techniques involving \textbf{domain-expert model annotation}, \textbf{LMM-as-judger} and \textbf{human-in-the-loop scrutiny}.

\item We present an efficient \textbf{prompt disentanglement} strategy that removes redundant textual guidance, enabling quality-scoring tasks to be completed solely from visual tokens. This avoids target confusion in training objectives caused by multi-task collective training. 


\item Building upon \textit{LLaVA-OneVision} and \textit{SlowFast}, we design an adaptive structure suitable for multi-modal inputs. After post-training on over $1M$ data samples, we obtain the \textbf{LLaVA-Assessor-GIGA}. This powerful all-in-one LMM achieves competitive performance on multiple tasks.
\end{itemize}

This study is based on our previous works~\cite{jia2025vqa2,jia2026scaling}, but with substantial differences.
Our earlier studies focused exclusively on video quality assessment~(VQA), whereas the present work extends the framework to a unified training paradigm that jointly addresses both image quality assessment~(IQA) and VQA. We also further address the \textbf{training-objective confusion} in multi-modal, multi-task joint training. We therefore systematically investigate the effects of different modality/task training configurations on the final training performance, providing a detailed analysis of how heterogeneous supervision signals influence model optimization and generalization. These important aspects have not been considered in our previous studies. In addition, we significantly expand the scale and scope of the experimental evaluation to provide a more comprehensive validation of the proposed model as well as detailed discussions of the effects of important experimental settings. 
\vspace{-8pt}
\section{Related Works}

\subsection{Visual Quality Assessment}
Fundamental visual quality assessment can be categorized into image Quality Assessment~(IQA) and video quality assessment~(VQA). IQA can be categorized based on the content type, including in-the-wild IQA~\cite{hosu2020koniq}, AI-generated content~(AIGC) IQA~\cite{agiqa3k}, panoramic IQA~\cite{sun2019mc360iqa}, etc. Early IQA methods rely on natural statistical information~(NSI)~\cite{ssim,brisque,niqe}. As distortions become more diverse and visual content grows increasingly complex, data-driven deep neural networks~(DNNs)~\cite{nima, dbcnn, hyperiqa,musiq,topiq} have gained prominence in the field.  VQA extends IQA by incorporating temporal factors, such as dynamic range, playback smoothness, and scene switching. It also covers various content, such as user-generated content~(UGC)~\cite{tu2021ugc}, professionally generated content~(PGC)~\cite{cheng2020screen,jia2024dsa,lu2022deep,yang2022blind}, and AIGC content driven by technological trends~\cite{zhang2025human,zhang2025q,chen2024gaia}. 

Studies have already leveraged vision-language models~(VLMs) to address specific tasks in visual quality assessment. Early works like \textit{CLIP-IQA+}~\cite{wang2023exploring}, \textit{LIQE}~\cite{liqe}, \textit{Explainable-VQA}~\cite{wu2023towards}, and \textit{Q-CLIP}~\cite{mi2025q} use \textit{CLIP}~\cite{radford2021learning}-based VLMs to integrate text information for quality scoring. \textit{Q-ALIGN}~\cite{wu2024q1} lays the foundation of the LMM-based quality scoring using the log-probability estimation strategy. \textit{LMM-PCQA}~\cite{zhang2024lmm} extends this to point cloud quality scoring, while \textit{LMM-VQA}~\cite{ge2024lmm} improves video quality scoring with refined LMM structures and a coarse-to-fine training pipeline. \textit{Compare2Score}~\cite{zhu2024adaptive} tackles subjective label scarcity using image pairs as pseudo-labels, and \textit{Fine-VQ}~\cite{duan2025finevq} creates a multi-dimensional subjective VQA dataset to improve fine-grained scoring. \textit{DeQA}~\cite{you2025teaching} explores leveraging subjective soft-label distributions to boost image quality scoring. \textit{Q-instruct}~\cite{wu2024q}, \textit{Aes-expert}~\cite{huang2024aesexpert}, and \textit{VQA\textsuperscript{2}}~\cite{jia2025vqa2} pioneer in training models with quality interpretation capabilities in image technical quality, image aesthetics, and video quality, respectively. \textit{Co-instruct}~\cite{wu2024towards} and \textit{DepictQA}~\cite{you2024depicting,you2024descriptive} focus on image pair quality comparison tasks. 
However, these works are limited to single modalities or tasks and lack a unified model, which motivates our development of the foundation LMM.

\begin{figure}
    \centering
    \includegraphics[width=\linewidth]{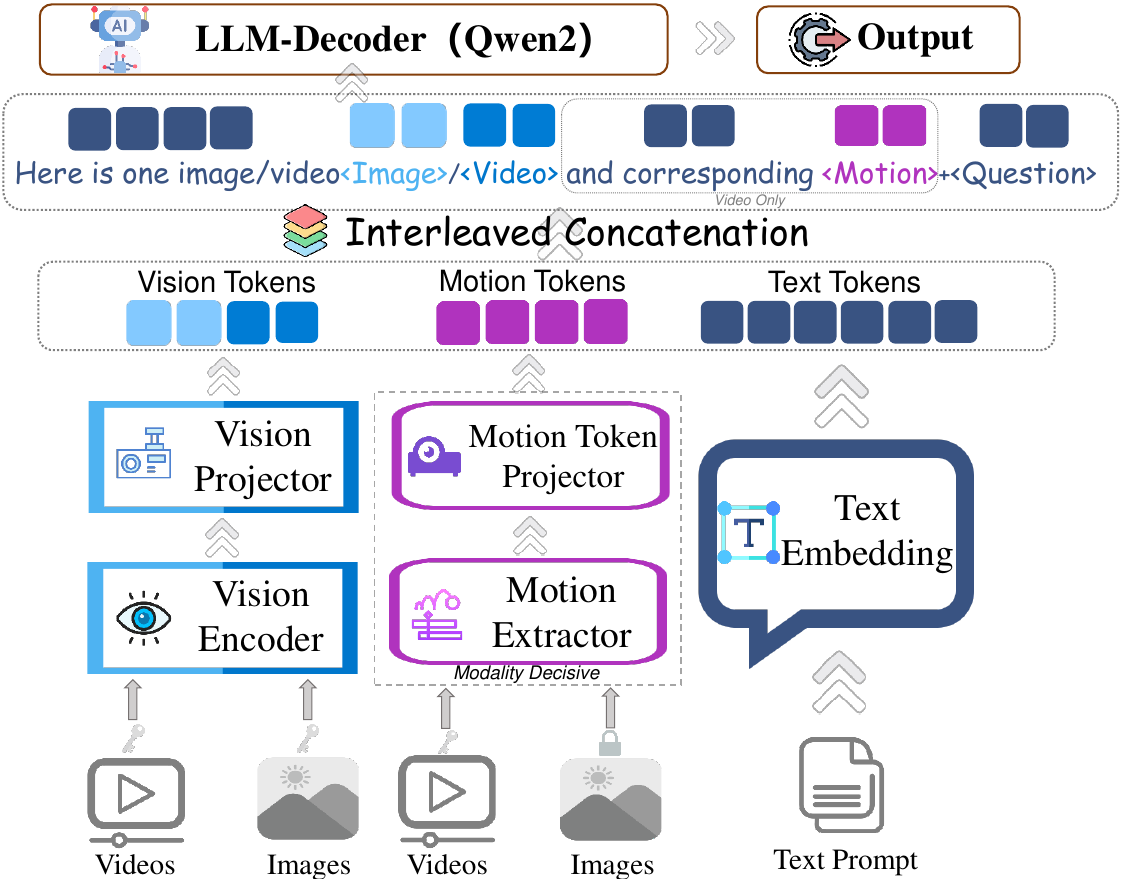}
    \vspace{-16pt}
    \caption{\textbf{The model structure} featured on the ``modal-selective" input strategy for different modalities. The vision token, motion token, and text token are concatenated in an interleaved manner that aligns with their semantic sequence. Specifically, the ``motion tokens" are applicable only for video inputs.}
    \label{fig:MODEL}
    \vspace{-0.2cm}
\end{figure}

\begin{table*}[h]
 \renewcommand\arraystretch{0.95}
\renewcommand\tabcolsep{16pt}
\belowrulesep=0pt\aboverulesep=0pt
\centering
\caption{
Details of the model structure configurations of the \textbf{LLava-Assessor-GIGA}. 
}
\vspace{-7pt}
\resizebox{\linewidth}{!}{
\begin{tabular}{l| c| c}
\toprule
\textbf{Model Structure} &  \textbf{Name} &  \textbf{More Information}  \\
\midrule
Vision Tower & \textit{SigLIP-SO400m} &\textit{Parameter size=}$397.75$M, \textit{Tokens per keyframe=$196$}  \\
Vision Projector & \textit{2-layers MLP+GeLU}&\textit{Parameter size}=$16.98$M \\
Motion Extractor& \textit{SlowFast-R50} &\textit{Use the fast-path feature, token numbers are the same as the frame number}  \\
Motion Projector& \textit{2-layers MLP+GeLU}&\textit{Parameter size}=$13.77$M, the same with the structure of vision projector\\
LLM init. & \textit{Qwen-2~(7B)} &Decoder-only model, \textit{parameter size}=$7660.56$M \\

\bottomrule
\end{tabular}
}
\label{tab:modelstruc}
\end{table*}
\begin{table}[!t]\tiny
    \centering
\renewcommand\arraystretch{1.1}
\renewcommand\tabcolsep{0.25pt}
\belowrulesep=0pt\aboverulesep=0pt
    \caption{Statistical summary of the training MIDB for our \textit{LLaVA-Assessor-GIGA}.}
\vspace{-5pt}
   \resizebox{0.97\linewidth}{!}{\begin{tabular}{c|c|c|c|c}
    \hline
    \textbf{Type} &\textbf{Categories} &\textbf{\# Instruction Pairs} &\textbf{Source Dataset} &\textbf{\# Video/Image} \\ \hdashline
    \multirow{2}{*}{\textit{Videos~(human)}}&\multirow{1}{*}{Scoring}&
     28,056& LSVQ &28,056 \\
     &Interpreting& 115,124 & LSVQ &15,500\\ \cdashline{1-5}
     \multirow{3}{*}{\textit{Videos~(machine)}}&\multirow{1}{*}{Technical}&
     230,000& Panda-70m&55,128 \\
     
     &In-context& 127,000 & Panda-70m &22,927\\ 
     &Aesthetic& 45,000 & Panda-70m &7,728\\ \cdashline{1-5}
     \multirow{3}{*}{\textit{Image}}&\multirow{1}{*}{Scoring}&
     13,000& Q-ALIGN-DB&13,000 \\
     &Technical& 200,000 & Q-Pathway-200K &18,973 \\ 
     &Aesthetic& 409,000 & AesMMIT &21,904 \\
    \hline
\end{tabular}
 }
\label{tab:datainformation}
\end{table}
\vspace{-5pt}
\subsection{MIDBs for Visual Quality Assessment}
\label{MIDB}
Many approaches have been proposed for constructing MIDBs for LMM training, which can be broadly grouped into three categories. The first, \textit{human-as-perceiver} methods~\cite{wu2024q,jia2025vqa2,you2024depicting,huang2024aesexpert,zhou2024uniaa}, rely on human experts to directly perceive and annotate visual signals, after which LLMs are often used to refine or rewrite for improved textual diversity. Although such annotations are closely aligned with human perception, this paradigm is costly in terms of labor and time, and is also susceptible to systematic subjective biases, thereby limiting the scalability of MIDB. The second category, \textit{general-LMM-as-perceiver-and-annotator} methods~\cite{you2024descriptive}, employs general-purpose LMMs, such as \textit{GPT-4o}~\cite{achiam2023gpt} and \textit{Gemini}~\cite{team2024gemini}, to perceive and annotate visual quality. However, the resulting annotations are often constrained by the limited domain-specific capability of teacher models that are not explicitly optimized for visual quality assessment. The third category, \textit{knowledge injection} methods~\cite{wu2024q1,chen2024grounding,wu2024towards,chen2024q}, enhances existing MIDBs by injecting task-specific knowledge and further reformulating the annotations with LLMs for downstream tasks. Nevertheless, the scalability of this strategy remains largely bounded by the size and scope of the source datasets.

Despite their differences, existing MIDB construction paradigms share several limitations. Most datasets primarily focus on either technical quality~\cite{wu2024q,jia2025vqa2,you2024depicting,you2024descriptive,wu2024q1,wu2024towards} or aesthetic quality~\cite{huang2024aesexpert,zhou2024uniaa}, resulting in limited versatility across multi-task and multi-modal scenarios. Moreover, the majority of existing MIDBs~\cite{wu2024q,jia2025vqa2,you2024depicting,huang2024aesexpert,zhou2024uniaa,you2024descriptive,wu2024q1,wu2024towards} provide only coarse-grained descriptions of image or video quality, while lacking fine-grained spatiotemporal annotations and evaluations. These limitations motivate the development of our foundation LMM for comprehensive visual quality assessment.

\section{The LLaVA-Assessor}

Intending to achieve superior \textbf{versatility} and \textbf{efficiency}, we have meticulously designed the model architecture, the training data construction pipeline, and the training methodology. In this section, we will systematically elaborate on each of these research components.


\subsection{Model Structure and Crucial Settings}
\subsubsection{Model Structure Design}
Our model structure is depicted in Fig.~\ref{fig:MODEL} and Tab.~\ref{tab:modelstruc}.
We adopt \textit{LLaVA-OneVision-Chat-7B}~\cite{li2024llava}. This model demonstrates superior performance across various advanced visual question answering benchmarks and is preferred in our previous studies. The core components of the model consist of: a vision tower built upon SigLIP~\cite{zhai2023sigmoid}, responsible for extracting vision tokens from the keyframe sequence; a vision projection module formed by fully connected layers designed for dimension transformation; and the \textit{Qwen-2}~\cite{team2024qwen2} paired with its tokenizer, which serves as the LLM decoder and handles text embedding.
It is observed that numerous LMMs~\cite{zhang2023video, li2024llava,ye2024mplug1,chen2024far} achieve strong performance on complex video-centric high-level tasks by solely leveraging sparsely sampled keyframe sequences. This phenomenon can be attributed to the inherent temporal redundancy of semantic information within videos, where consecutive frames often share highly similar high-level features. Consequently, significant temporal transitions or relationships spanning longer durations can still be effectively represented through sparse samplings.
Nevertheless, such temporal redundancy largely diminishes in the context of visual quality assessment. Subtle temporal artifacts such as stuttering or jitter, which occur over very short periods, can drastically impair perceived video quality. These distortions are tightly linked to immediate frame-to-frame variations and thus cannot be captured if the keyframes are sampled sparsely. 

\begin{figure*}
    \centering
    \includegraphics[width=\linewidth]{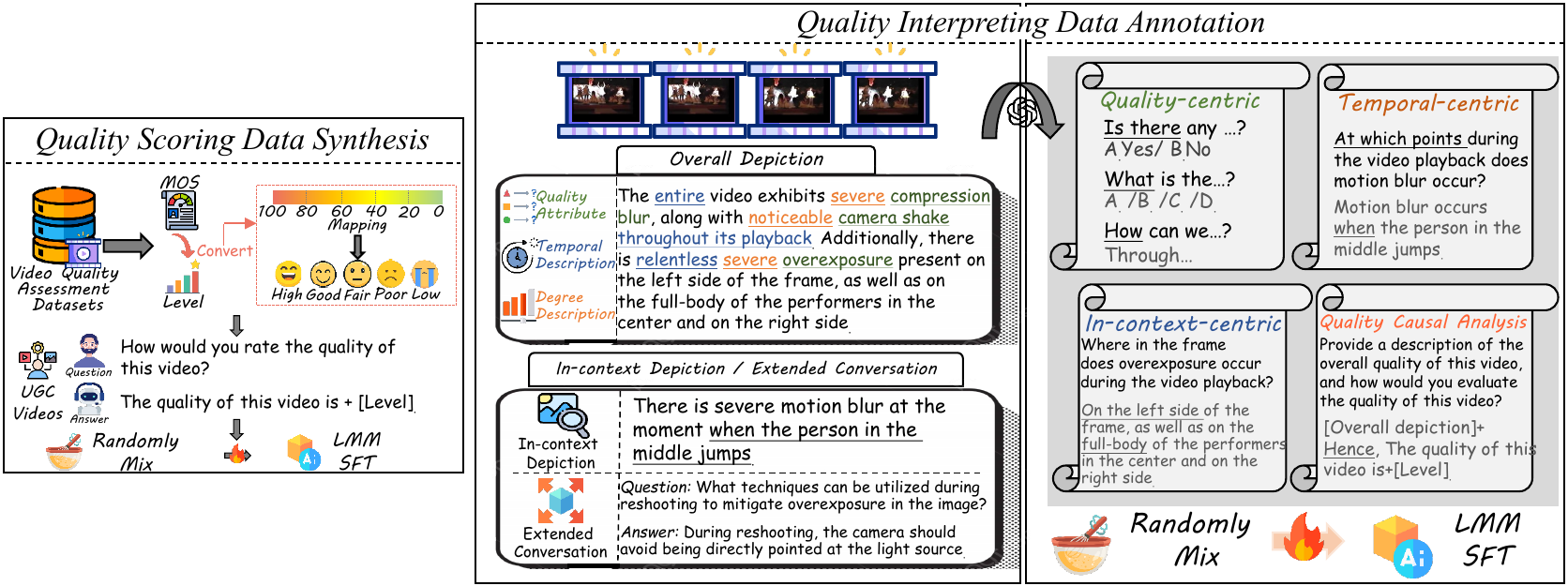}
     \vspace{-20pt}
    \caption{\textbf{The human-annotated data construction pipeline}. The quality scoring data~(left part of the figure) primarily derives from existing video quality assessment datasets and is converted based on quality level. The quality interpretation data~(on the right part) is annotated via human experts based on three aspects: ``quality factor", ``temporal description", and ``degree description". Additionally, in-context descriptions or extended conversations are provided. These annotations are then rewritten by \textit{GPT} into multiple types of question-answer pairs suitable for SFT training.
}
    \label{fig:PIPELINE-SFT}
   
\end{figure*}
Hence, we advocate integrating a dedicated video motion extraction module that captures temporal dynamics by analyzing adjacent frames from densely sampled video inputs. We employ the \textit{SlowFast-R50}~\cite{feichtenhofer2019slowfast} to extract motion features, processing the entire video sequence following spatial pre-processing steps. To synchronize the motion token number with the total number of video frames, we configure the slow pathway’s sampling interval to $\tau=4$. Then, we set the ratio of slow to fast sampling rates as $\alpha=4$. Only the fast pathway outputs are utilized to generate the motion token sequence. Subsequently, these motion tokens are projected through a motion projector, which shares the architectural design of the vision projector in \textit{LLaVA-OneVision}, thereby aligning the motion token dimensions with other tokens.


\subsubsection{Crucial Setting-1: Interleaved Token Combination}
The \textit{LLaVA-Assessor} supports the \textbf{interleaved} token input format, wherein special tokens representing visual and motion features are interspersed within the textual content \textbf{according to their semantic order}. During both training and inference, these placeholders are subsequently substituted with their corresponding extracted tokens, thus enabling seamless integration of multi-modal information.

\subsubsection{Crucial Setting-2: Video-Centric Data Collection}
We design the subsequently illustrated  training MIDB construction pipeline with a \textbf{video-centric setting}. We consider its rationality from the two aspects below. 
\begin{enumerate}
\item 
First, the evaluation quality factors for VQA incorporate almost every spatial quality factor inherent in IQA. Accordingly, the MIDB tailored for video content can comprehensively encompass these fundamental visual quality dimensions. Subsequent experiments in \ref{ablation} also demonstrate that, in quality assessment tasks, the generalization from image to video is easier to achieve than that from video to image.
\item 
Meanwhile, due to the significant gaps in both quantity and completeness of quality description MIDBs for video compared to those for image, this design serves as an effective supplement to this data deficiency.
\end{enumerate}
The final training MIDB consists of the combined video and image datasets. The statistics for all the training MIDB are presented in Tab.~\ref{tab:datainformation}.

\vspace{-10pt}
\subsection{Human-annotated Data Construction}
\label{Human-data}
\subsubsection{Video Quality Scoring Data Preparation}
We use the LSVQ~(train)~\cite{ying2021patch} as the source dataset for this task.  We normalize the MOSs in the dataset to the $[0,100]$ range. During the training phase, we map continuous scores to discrete rating levels. Specifically, we divide the range between the highest MOS~($\mathrm{M}$) and the lowest MOS~($\mathrm{m}$) into $5$ equal intervals. MOSs falling within each interval are then assigned to the corresponding quality level:
\begin{equation}
    {L(s)} \! = \! l_i \text{  if } \text{m}\! + \!\frac{i-1}{5} \!\times \!\mathrm{(M-m)} \!< \!s \! \leq \! \mathrm{m} \!+\! \frac{i}{5} \!\times\! \mathrm{(M-m)},
\end{equation}
where \{$l_i|_{i=1}^{5}\}=\{\textit{low, poor, fair, good, high}\}$ are the standard quality levels. This approach minimizes the impact of inconsistent quality distributions across datasets.
The format of the instruction pairs for this task is as follows~(we discard the prompt-disentanglement trick explained in Sec.~\ref{PD} here for better illustration):

\noindent\textit{USER: How would you evaluate the quality of this video?} \\
\noindent\textit{ASSISTANT: The quality of this video is \textbf{[Level]}.}

\subsubsection{Video Quality Interpreting Data Annotation}
 The core component of the human-annotated MIDB is the quality-interpretation instruction subset. We select $14,005$ videos from LSVQ~(train) and LSVQ~(1080p)~\cite{ying2021patch}, together with $998$ videos from the AIGC video dataset Videofeedback~\cite{he2024videoscore}. This subset mainly targets low-level technical quality, while also containing a small amount of data related to video aesthetic assessment~(VAA) and AIGC analysis.

For each video, the human annotation process consists of two stages. First, annotators provide a comprehensive \textbf{overall quality depiction} centered on representative quality factors. Each depiction includes several quality factors, each described by three elements: {\textit{Quality Factor}}$+${\textit{Degree}}$+${\textit{Temporal Description}}. The detailed pipeline and key annotation  factors are shown in Fig.~\ref{fig:PIPELINE-SFT} and Tab.~\ref{tab:qualityfactorsexp}, respectively. Second, annotators provide a brief depiction of in-context local temporal or spatial quality. When no salient local spatiotemporal quality pattern is observed, this depiction can be replaced with extended conversations, such as designing a Q\&A pair about the possible causes of certain distortions or proposing feasible solutions for video quality enhancement.

This annotation strategy reflects the key observation that viewers perceive video quality by attending to \textbf{prominent quality features}. By explicitly incorporating temporal analysis, the framework substantially improves the model's ability to answer questions about temporal aspects of video quality. Human annotators are instructed to select quality factors that best represent the overall video quality, guided by the quality level of each video. For medium-quality videos, annotators are further encouraged to capture a broader range of quality issues, covering both positive and negative elements, thereby improving the utility of such samples.
For the annotated overall depictions mentioned before, to fully utilize the human-annotated data during training, we manually formulate them into the form of quality causal analysis with the specific format shown in Fig.~\ref{fig:PIPELINE-SFT}.
\subsubsection{The \textit{LLaVA-Assessor-SFT}}
We apply the supervised-finetuning~(SFT) using the previously obtained human-annotated MIDB to train the \textit{LLaVA-Assessor-SFT~(7B)} upon the base model. This serves as the  primary machine annotator for subsequent dataset expansion in Sec.~\ref{machine}. We sequentially input the quality scoring data and quality understanding data, training the model for one epoch only. 
The training process uses a typical cross-entropy~(CE) loss:
\begin{equation}
\mathcal{L} = - \frac{1}{L}\sum_{\ell=0}^{L-1} \log p(z_{\ell} | \mathbf{Z_{all}}, z_{<=\ell}),
\end{equation}
where $\mathbf{z_{\ell}}$ represents the output probability of the ${\ell}$-th target token, $\mathbf{Z_{all}}$ denotes the interleaved input tokens, and $L$ represents the length of the target sequence.
\begin{table*}[h]\small
\renewcommand\arraystretch{0.85}
\renewcommand\tabcolsep{15pt}
\caption{\textbf{The representative quality factors examples for human annotation}. The quality factors in \textbf{bold} are those that must be considered in most videos. The quality factors in \textit{italic} font are the ones that can only be noticed in a small number of videos. The remaining factors should be considered depending on the content and quality level of the specific video.}
 \vspace{-8pt}
\centering
\begin{tabular}{p{6cm}|p{10cm}}
\hline
\textbf{Quality Factors} & \textbf{Standard Annotation Examples} \\
\hline
Sharpness & Sharp, relatively clear, relatively fuzzy, very blurry \\ 
\hdashline
\textit{Focus}& In-focus, out-of-focus \\
\hdashline
Noise & Noiseless, a small amount of noise, severe noise densely distributed \\
\hdashline
Motion Blur & Clear-motion, blur-motion \\
\hdashline
Flicker/Camera Shake & Stable, small amplitude shake, shaky \\
\hdashline
Exposure~(subjective to light) & Well-exposed, underexposure/overexposure \\
 \hdashline
Compression Artifact & Almost with no compression, with noticeable compression artifacts, severe compression blur, significant blockiness, serious loss of edge/texture details \\
\hdashline
Fluency & Smooth playback~(fluent),~(short-term) stutter/playback jitter \\
\hdashline
\textit{Contents} & Logically presented content, completely illogical \\
\hdashline
\textit{Composition} & Well-organized composition, poorly arranged frames, chaotic  \\
\hdashline
Color & Vibrant, natural, single, faded, unnatural\\
\hdashline
Light & Natural, soft, high contrast, uneven, dark, underexposed, overexposed \\
\hdashline
Camera Trajectory & Coherent and stable, consistent, poor, shaky, lack consistency and coordination \\
\hdashline
\textit{Quality Switch} & Quality maintains at a consistent level, sharpness drops/rises at a certain moment / undergoes multiple violent changes within a certain time interval \\
\hline
\textbf{Degree Description} & \textbf{Standard Annotation Examples} \\
\hline
        Very/Extremely Severe~(used primarily for quality levels ``Poor" and ``Low") & The presence of such extremely severe distortions significantly degrades the viewer's quality of experience. The primary presence of such distortion leads to a very negative viewing experience, causing the overall experience to deteriorate significantly/important details in certain scenes are completely destroyed with such distortions. \\
        \hdashline
        Relatively/Quite Severe~(used primarily for quality levels ``Fair", ``Poor", and ``Low")  & The presence of such distortion moderately degrades the viewer’s overall viewing experience. This distortion is relatively prominent, annoying, and difficult to ignore. \\
        \hdashline
        Mild/Relatively Slight/Merely Noticeable~(can be used in videos with all levels)  & The presence of such mild distortion has a relatively light but nontrivial impact on the viewer’s experience. \\
        \hdashline
        Good~(used primarily for quality levels ``High" and ``Good") & This quality factor is relatively good, and does not affect the viewing experience.\\
        \hdashline
        Excellent/Flawless~(used only for the quality level ``High") & This degree description is recommended when the quality factor is outstanding and almost flawless, while significantly enhancing the viewer’s experience. \\
        \hline
        \textbf{Temporal Description} & \textbf{Standard Annotation Examples} \\
        \hline
         Overall Description & Appears throughout the entire playback / high frequency / multiple occurrences. \\
        \hdashline
        {In-context Temporal Description} &\textit{When}  Someone is doing something / \textit{When} event A occurs / \\ 
                                     & Camera switches \textit{from} Scene A \textit{to} Scene B  \\ 
\hdashline
{Direct Temporal Description} & At the very beginning of the video / At the $3$-rd second mark / \\ 
                                      &When the video is about to end / From $3$-$7$ seconds in the video \\ 
        \hline
\end{tabular}
\label{tab:qualityfactorsexp}
\end{table*}

\subsection{Scaling-up the Data Via Machine-dominated Annotation}
\label{machine}
\subsubsection{Candidate video pool construction}
We select $100,000$ videos from a large-scale UGC video pool Panda-70m~\cite{chen2024panda} to serve as the source video set. We impose the constraint that the length of all candidate videos must be in the range of $[3,15)$ seconds. We then utilize $4$ state-of-the-art objective video quality scoring methods~\cite{wu2022fast,wu2023exploring,wu2024q1,sun2024analysis} to label the objective quality for the candidates. To ensure consistency in the scale of scores across different rating methods, the scores are first normalized to the range of $[0,100)$ and then averaged for each video to determine its objective quality label.
\begin{figure*}
    \centering
    \includegraphics[width=0.98\linewidth]{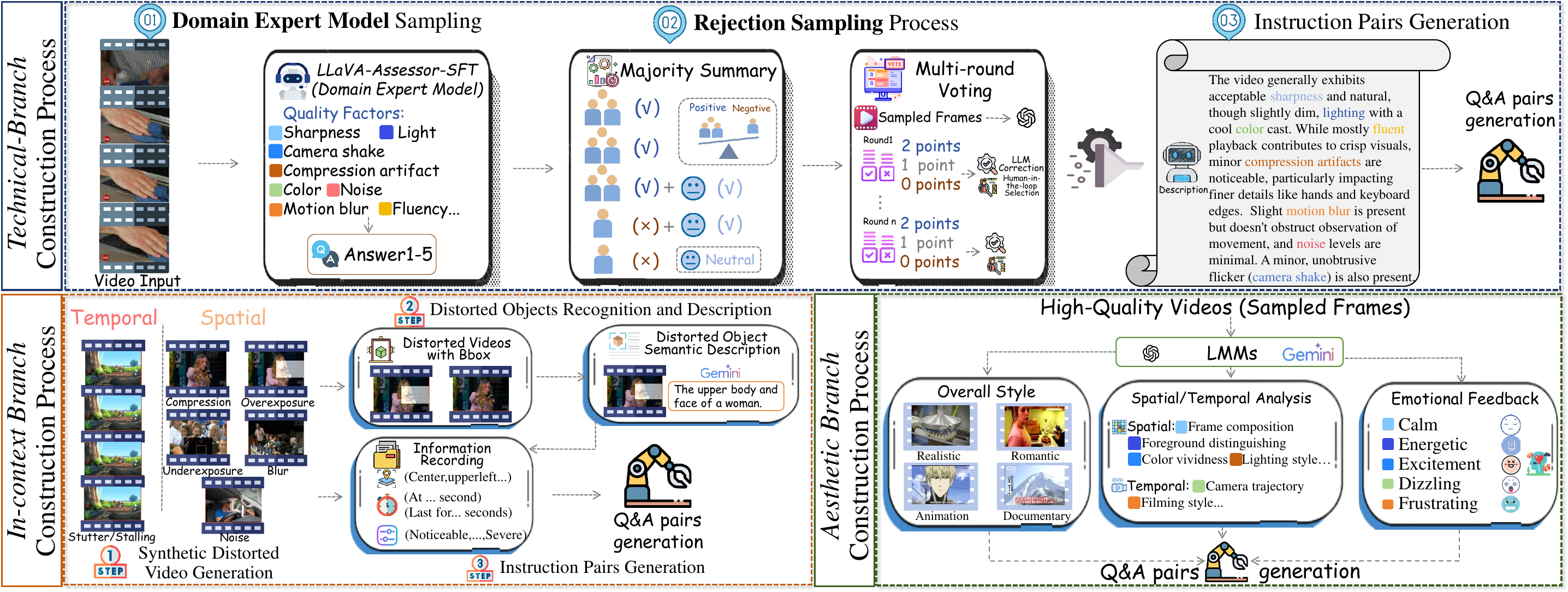}
     \vspace{-8pt}
    \caption{\textbf{The machine-dominated annotation process}. This process includes three branches: the \textbf{technical branch}, which uses a domain-specific model fine-tuned on the previous human-annotated data as the primary annotator, and incorporates human-in-the-loop and LMM-as-judger strategies for rejection sampling to construct data; the \textbf{in-context branch}, which uses LMMs in conjunction with synthesized local distortion data to build the dataset; and the \textbf{aesthetic branch}, which selects appropriate aesthetic evaluation tasks for LMM annotation, utilizing carefully designed prompts for LMM annotation.
}
    \label{fig:PIPELINE-RFT}
\end{figure*}
\subsubsection{Technical Quality Data Expansion}
The \textbf{technical quality} is the primary quality concern of the in-the-wild VQA. Here, we randomly select $55,128$ videos from the candidate video pool, ensuring content and quality diversity. Each video is annotated across $8$ quality factors: \textbf{sharpness}, \textbf{light}, \textbf{compression}, \textbf{color}, \textbf{noise}, \textbf{fluency}, \textbf{motion blur}, and \textbf{camera shake}. The annotation process is demonstrated in Fig.~\ref{fig:PIPELINE-RFT}.

We contend that using the general LMMs as a \textbf{``judger''} provides a more efficient and practical method for distillation than directly using them as video quality annotators. This technique proves more effective in steering general LMMs toward generating more accurate and reliable responses.
Hence, we introduce a novel approach for annotating each quality factor based on \textbf{rejection sampling}. The process begins with conducting $N=5$ multiple samplings from the outputs of the domain expert model~(the \textit{LLaVA-Assessor-SFT}), which together form the \textbf{suggested distribution}. We then employ a heuristic approach that maximizes the utilization of all available sample information. This approach filters and summarizes the samples obtained during the sampling process using a reasoning LLM, the \textit{Openai-o1}~\cite{openai2024o1}. The information in the responses obtained before is categorized into $3$ types: 
\begin{enumerate}
    \item \textbf{Positive information} is that it has similar meanings in the responses and appears in at least $3$ of the $5$ responses. These answers are then merged into the summary.
    \item \textbf{Negative information}, which contradicts the meaning of the positive information, is excluded from the summary.
    \item If any responses contain additional \textbf{neutral} information, it should be included in the summary.
\end{enumerate}
For example, for the quality factor ``\textbf{sharpness}" and the question \textit{``How is the sharpness of this video?"}, if the $5$ responses are: \textit{``Poor"}~(positive),  \textit{``The sharpness is relatively poor"}~(positive), \textit{``Poor, with degraded facial details"}~(positive with neutral), \textit{``Good, however, the facial details are slightly lost"}~(negative with neutral), \textit{``Excellent"}~(negative).
The summarized response would be:  
\textit{``The sharpness is poor with degraded human facial details."}

Subsequently, a voting mechanism, utilizing general LMMs, is then employed to guide the next steps of the process. In this context, general LMMs serve as \textbf{judgers} rather than annotators. 
We then utilize \textit{GPT-4o} to determine the appropriate post-processing strategy. The process begins by inputting the keyframe sequence, sampled at $1$fps, along with the prior summaries of each quality factor~(since the keyframe sequences cannot capture stuttering, if any of the $5$ original annotations indicate the presence of stuttering in the video, this video segment is directly assigned to human experts for evaluation, thereby bypassing the \textit{GPT} voting process).
Following this, we instruct \textit{GPT} to perform $3$ rounds of voting for each quality factor, evaluating the accuracy and relevance of the provided summaries, and assigning a score from the set $(2,1,0)$. Based on these votes, the post-processing method is determined and applied to the summaries: 
\begin{itemize}
    \item If all voting rounds yield a score of $2$, the summary is passed directly.
    \item If any voting round results in a score of $1$, the summary from that round is sent to \textit{GPT} for revision, and the revised output becomes the input for the subsequent voting round.
    \item In cases where a voting of $0$ appears in any of the rounds, \textit{GPT} is required to provide its correct summary, which, along with the original summary, is stored in a \textbf{cache}. These cached rollouts are then presented to human experts for selection~(Consequently, each video segment’s cache may contain up to $4$ candidate summaries: the original input summary plus up to $3$ GPT-generated revisions corresponding to each voting round).
\end{itemize}
For the human-in-the-loop selection process, human experts are required to choose \textbf{only one} description they consider most accurate from the cached summaries corresponding to each video. If making a definitive choice is difficult, experts can manually provide their comments or feedback. After all the voting and postprocessing, we employ \textit{o1} to consolidate the annotations across all quality factors, producing a comprehensive video-level quality summary. Utilizing this summary, the model is prompted to generate three distinct types of question-answer~(Q\&A) pairs focused on the quality factors: binary-choice questions, multiple-choice questions, and open-ended questions. To maximize the informational value of the quality summary, an additional quality summarization is also included for each video: 

\noindent\textit{USER: Please describe the overall quality of this video, evaluating as many quality factors as possible.} \\
\noindent\textit{ASSISTANT: The corresponding video-level quality summary.}


\subsubsection{In-context Data Injection}
The in-context data injection procedure aims to enhance the model’s capability to detect and characterize fine-grained local spatiotemporal quality distortions within videos. To reduce the confounding effects of pre-existing distortions, we select $6,500$ source videos from the candidate pool, each having objective quality scores exceeding $70$. The annotation  procedure is illustrated in the \textbf{lower left} section of Fig.~\ref{fig:PIPELINE-RFT}.

We artificially introduce localized spatiotemporal distortions to the videos. Spatial distortions include \textbf{overexposure}, \textbf{underexposure}, \textbf{blur}, \textbf{compression artifacts}, and \textbf{noise}, whereas temporal degradation specifically involves \textbf{video stuttering}.

Spatial distortions are applied randomly within a rectangular region covering one-quarter of the frame area. The starting time point and interval of these spatial distortions are assigned randomly as an integer duration between $1$ and $3$ seconds. Distortion intensity is classified into three tiers: \textbf{noticeable}, \textbf{relatively severe}, and \textbf{severe}. Each source video receives exactly one instance of each of the five spatial distortion types, applied sequentially. To improve the LMM’s ability to semantically interpret and describe the spatial distortion regions, we overlay a highlighted bounding box~(bbox) around the distorted rectangle in the generated videos. The keyframes containing these bboxes are input into the \textit{Gemini-1.5-Pro}~\cite{reid2024gemini15}, which identifies and describes primary semantic objects within the bbox under strict criteria: \textit{an object is considered valid for annotation only if it is entirely enclosed within the bbox, constitutes the main focus within this area, contrasts distinctly with its surroundings, and remains within the bbox for the entire duration of the distortion.} Finally, we record detailed metadata for each distortion event, including \textbf{start time, duration, type, intensity, and spatial location}.

For temporal distortion, we simulate video stuttering by excising $1$  second immediately following a randomly chosen integer timestamp and duplicating the preceding frames for an additional second, thereby producing a \textbf{frame freeze} effect. Each video is embedded with between $1$ and $3$ such stutter events, randomly distributed. Furthermore, generated videos in which the synthetic spatiotemporal distortions are not perceptually salient are manually excluded.

Subsequently, we generate $5$ instructional Q\&A pairs per distorted video, concentrating exclusively on the spatiotemporal local distortions. If the distortion involves annotated semantic objects, at least one Q\&A pair must pertain to those objects. We also incorporate an additional Q\&A format—\textit{cloze completion}—specifically targeting local distortion information.
Consistent with prior steps, each video includes a summary Q\&A pair: 

\noindent\textit{USER: Please describe the information of the spatiotemporal local distortions of the video.} \\
\noindent\textit{ASSISTANT: The recorded and summarized local distortion details of the video.}

\subsubsection{Mixing with Video Aesthetic Quality Annotations}
Proprietary LMMs excel at high-level tasks, thus are suitable for high-level MIDB scaling-up, so we design another video aesthetic branch, completely annotated by a general LMM, to broaden the diversity of the data. To ensure that technical distortions do not interfere with the extraction of aesthetic features, we carefully select $7,728$ videos exhibiting objective quality scores above $70$. Keyframes sampled from these videos are then directly fed into \textit{GPT-4o} to annotate aesthetic quality. 


The annotations are performed across three dimensions, as depicted in the \textbf{lower right} section of Fig.~\ref{fig:PIPELINE-RFT}:
\begin{enumerate}
\item \textbf{Aesthetic Style}: Give a brief description of the video's overall style~(from its type or main content style).
\item \textbf{Spatiotemporal Aesthetic Analysis}: Implement a comprehensive evaluation of aesthetic elements from both spatial and temporal perspectives.
\item \textbf{Emotional Impact}: Deliver a description of the emotional responses the video is likely to elicit from viewers.
\end{enumerate}
The rationality behind these settings is: the  \textit{overall style} and \textit{emotional feedback} annotation tasks align with what \textit{GPT-4o} are capable of. For \textit{aesthetic spatiotemporal analysis task}, we design detailed instructions to ensure general LMMs avoid involving complex low-level distortion analysis, thereby maintaining annotation accuracy.

The machine-generated annotations for each video are then consolidated into a summary, from which $6$ instruction Q\&A pairs are subsequently derived. Each annotated video is also accompanied by a summarization Q\&A pair:

\noindent\textit{USER: Please describe the aesthetic effects.} \\
\noindent\textit{ASSISTANT: The machine-annotated aesthetic information.}

\vspace{-10pt}
\subsection{Collective Model Training}
\label{setting}
\subsubsection{Data Combination}
Upon the above constructed video-centric MIDB, we firstly combine a large set of existing IQA MIDBs for mix training, including the quality scoring MIDB \textit{Q-ALIGN-onealign-DB}~\cite{wu2024q1}, the technical quality interpreting MIDB \textit{Q-Pathway-200K}~\cite{wu2024q}, and the image aesthetic interpreting MIDB \textit{AesMMIT}~\cite{huang2024aesexpert}. Furthermore, to enhance the efficiency of dataset utilization, we also combine the human-annotated MIDB in Sec.~\ref{Human-data}. During training, video data samples are assigned the ``video" modality label, while image data entries are assigned the ``image" modality label. The modality label functions as a \textbf{gating mechanism}, determining whether the input bypasses the \textit{SlowFast} and thus enabling collective multi-modal training.
\begin{figure}
    \centering
    \includegraphics[width=\linewidth]{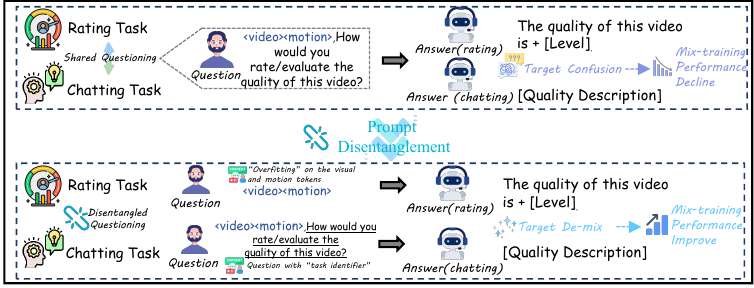}
    \vspace{-20pt}
    \caption{The illustration of the \textbf{prompt disentanglement} strategy~(we use VQA tasks as an example). The core lies in minimizing the similarity of guiding prompts across different tasks during SFT to avoid training target confusion.}
   
    \label{fig:PD}
    
\end{figure}
\subsubsection{The Prompt Disentanglement Strategy:}
\label{PD}
The prompt disentanglement strategy is illustrated in Fig.~\ref{fig:PD}. A key challenge in multi-task mixed training lies in \textbf{training-objective confusion}, which causes noticeable performance degradation, particularly on quality interpretation, as evidenced by the ablation study~(Sec.~\ref{ablation}, Tab.~\ref{tab:ablation}). For example, in the quality scoring task, Q\&A pairs usually contain prompts such as \textit{How would you rate the quality of this video?''}, whose responses are restricted to discrete quality levels or scores. By contrast, in the quality interpretation task, many prompts adopt a semantically similar form, such as \textit{How would you evaluate~(or describe) the quality of this video?''}, yet their expected responses are comprehensive textual descriptions of perceptual quality. Although these two prompt types are close in semantic expression, their supervision targets differ substantially: the former focuses on score level prediction, whereas the latter requires detailed quality reasoning and descriptive generation. When these samples are jointly optimized within the same training batches, the semantic ambiguity of prompts and the discrepancy in target formats may introduce conflicting supervision signals, thereby disturbing the optimization trajectory and degrading multi-task learning performance.

To address this issue, we observe that the quality scoring task essentially exploits the LLM decoder as a \textbf{numeric regressor}. In this task, the desired outputs belong to a compact and predefined label space, typically corresponding to predicted quality scores or quality levels. Therefore, semantic guiding prompts are not strictly necessary for steering the textual generation. Instead, the input can be simplified to visual tokens alone, without accompanying textual prompts. Through SFT, the model learns to directly map visual representations to the corresponding scores, thereby avoiding unnecessary semantic overlap with interpretation-oriented prompts. In contrast, the quality interpretation task still adopts diverse semantic prompts during training, so as to fully exploit the linguistic diversity of the training MIDB and encourage fine-grained and context-aware textual outputs. We refer to this strategy as ``\textbf{prompt disentanglement}''. Specifically, for sub-tasks with fixed output formats, guiding prompts are removed or simplified, encouraging the model to specialize in a distinct input--output pattern that does not interfere with other task prompts. This design effectively mitigates training-target ambiguity and improves the stability of multi-task collaborative learning.

Under this strategy, we randomly mix all previously constructed data and perform SFT using the standard cross-entropy loss, obtaining the final unified foundation LMM, \textbf{\textit{LLaVA-Assessor-GIGA~(7B)}}. By separating prompt forms according to task-specific output structures, the model can share visual-language representations across modalities and tasks while preserving distinct generation behaviors for scoring and interpretation. This enables a unified model to perform both quantitative quality prediction and qualitative quality reasoning within a coherent training framework.

\section{Experiments}
To rigorously validate and analyze the model’s versatility and efficiency, we devise meticulous experimental configurations and conduct comprehensive comparative experiments alongside detailed ablation studies and discussions.
\begin{table}[t]
 \renewcommand\arraystretch{0.85}
\renewcommand\tabcolsep{14pt}
\belowrulesep=0pt\aboverulesep=0pt
\centering
\caption{
Details of the hyperparameters for model training. The \textcolor{red}{red} /  \textcolor{blue}{blue} colors represent the hyper-parameters used in the \textcolor{red}{LLaVA-Assessor-SFT} and \textcolor{blue}{LLaVA-Assessor-GIGA} training, respectively. Other entries without color differentiation indicate that the hyperparameter remains consistent across both rounds of training.
}
\vspace{-7pt}
\resizebox{\linewidth}{!}{
\begin{tabular}{l| c}
\toprule
\textbf{Hyper-Parameters} &  \textbf{Value}\\
\midrule
Image Input Resulution &Original\\ 
Keyframes Sampling Rate &$1 fps$\\   
Keyframes~(for videos) Resolution         & $384\times 384$\\

Frames~(for motion extraction) Resolution         & $192\times 192$\\
Batch Size& 8 \\
LR Max & \textcolor{red}{1e-5} / \textcolor{blue}{1e-6}  \\
LR Schedule & cosine decay \\
Warmup Epochs & $0.03$ \\
Weight Decay & $0$ \\
Gradient Accumulation Steps   & $\textcolor{red}{1}$ / $\textcolor{blue}{2}$ \\
Numerical Precision      & $\mathtt{bfloat16}$ \\
Epoch &\textcolor{red}{$1$} / \textcolor{blue}{$2$} \\
Optimizer & AdamW \\
Gradient Checkpointing &\checkmark \\
Deepspeed Stage & $\textcolor{red}{2}~/~\textcolor{blue}{3}$ \\
\bottomrule
\end{tabular}
}
\label{tab:hyperparam}
\end{table}
\vspace{-10pt}
\subsection{Experiment Configurations}
\subsubsection{System prompts design}
For different modalities and sub-tasks, we employ distinct system prompts during training. For the quality scoring task, under the prompt disentanglement setting, we do not apply any additional system prompts. 
For the quality interpretation task, we set different system prompts for images and videos to help the model better understand the specific tasks associated with each modality:

\noindent Image: \textit{You will receive an image \textbf{\textit{[image]}}. Please answer the following question based on the information provided.} 

\noindent Video~(general tasks): \textit{You will receive a keyframe sequence sampled at $1fps$ from a video of {length} seconds: \textbf{\textit{[video]}} with the keyframe sequence ordered in alignment with the video’s temporal order. In addition, you will receive a motion feature sequence that corresponds to the number of frames in the video: \textbf{\textit{[motion]}} {num of frames}. Please answer the following question based on the information provided.}

For the spatiotemporal fine-grained tasks~(in-context branch), which involve specific time-related questions, we introduce a standardized temporal representation to ensure consistency in the responses from different LMMs: 
\noindent \textit{When the video starts playing, this timepoint is denoted as ``1 second". When the 1-st second ends and the 2-nd second begins, this timepoint is marked as ``2 seconds', and so on.}
\begin{figure}
    \centering
    \includegraphics[width=\linewidth]{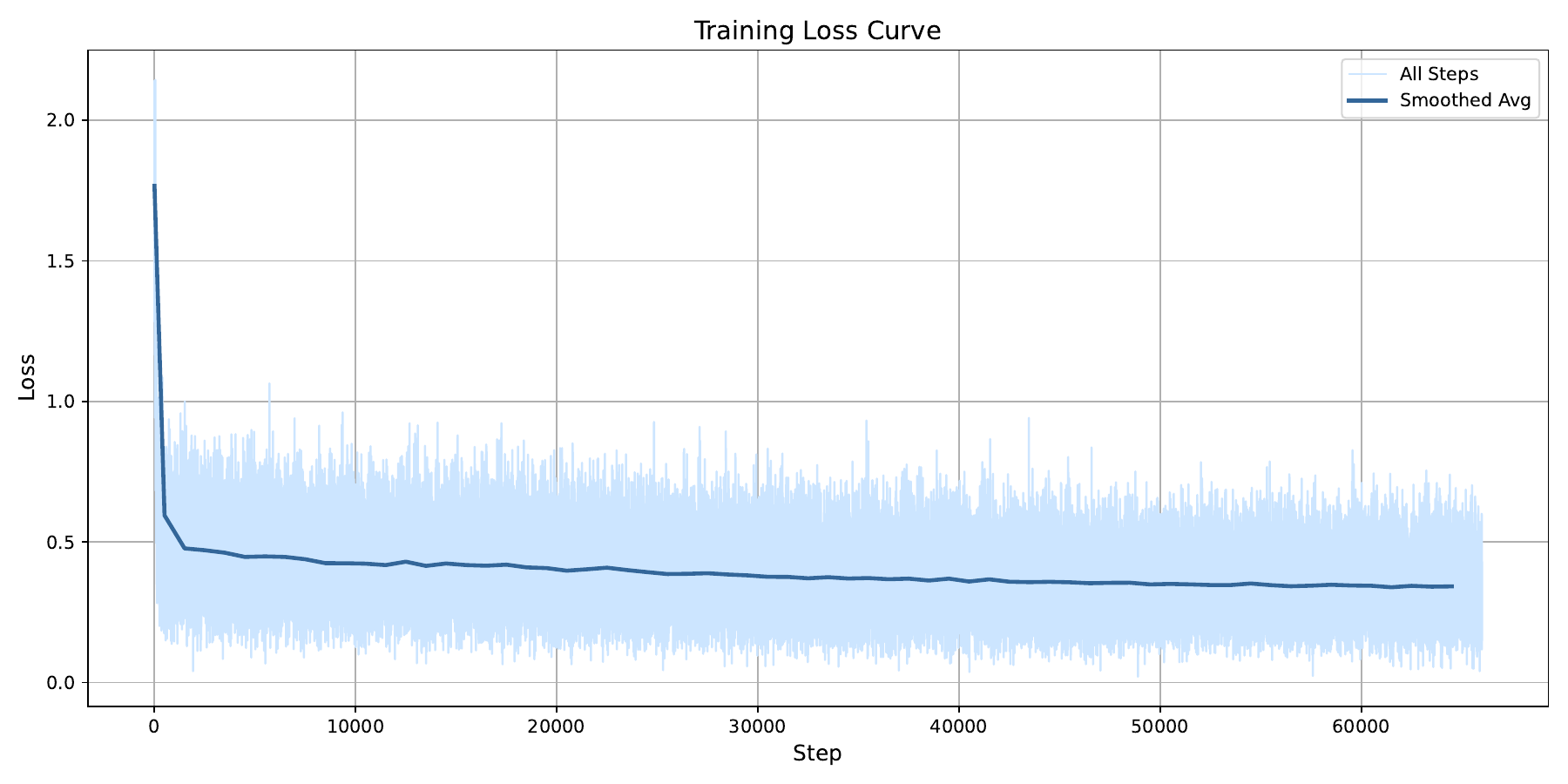}
    \vspace{-20pt}
    \caption{The training loss curve along with the training steps. It can be observed that the training loss drops quickly and then stabilizes.}
    \label{fig:trainingloss}
    
\end{figure}
\begin{table*}[t]\tiny
    \centering
    \renewcommand\arraystretch{1.05}
    \renewcommand\tabcolsep{4pt}
    \belowrulesep=0pt\aboverulesep=0pt

    \caption{Performance on video quality scoring tasks. Per column: highest in \textcolor{red}{red}, second in \textcolor{blue}{blue}, third \underline{underlined}.        [{$\spadesuit$}: Intra-dataset testing, $\diamondsuit$: Inter-dataset testing, \textit{Textit}: Out of domain test datasets]}
    \vspace{-8pt}
    \resizebox{\linewidth}{!}{%
    \begin{tabular}{l|ccccccccccccc}
    \hline
    \multicolumn{1}{l|}{\textbf{Datasets}}
      & \multicolumn{2}{c}{\textbf{LSVQ(1080p)~\cite{ying2021patch}\textsuperscript{$\spadesuit$}}}
      & \multicolumn{2}{c}{\textbf{LSVQ(test)~\cite{ying2021patch}\textsuperscript{$\spadesuit$}}}
      & \multicolumn{2}{c}{\textbf{LIVE-VQC~\cite{sinno2018large}\textsuperscript{$\diamondsuit$}}}
      & \multicolumn{2}{c}{\textbf{KoNViD-1K~\cite{hosu2017konstanz}\textsuperscript{$\diamondsuit$}}}
      & \multicolumn{2}{c}{\textbf{YT-UGC~\cite{wang2019youtube}\textsuperscript{$\diamondsuit$}}}
      & \multicolumn{2}{c}{\textbf{\textit{YT-Gaming}~\cite{yu2022subjective}\textsuperscript{$\diamondsuit$}}}
      & \multirow{2}{*}{\textbf{Avg.}} \\
    \cline{1-13}
      \textbf{Models} & SRCC & PLCC
      & SRCC & PLCC
      & SRCC & PLCC
      & SRCC & PLCC
      & SRCC & PLCC
      & SRCC & PLCC & \\
    \cline{1-14}
    Simple-VQA~(ACM.MM 2022)~~\cite{sun2022deep}
      & 0.760 & 0.805
      & 0.870 & 0.868
      & 0.755 & 0.793
      & 0.862 & 0.859
      & 0.826 & 0.821
      & 0.686 & \textcolor{blue}{0.746}
      & 0.804 \\
    BVQA~(TCSVT 2022)~~\cite{li2022blindly}
      & 0.747 & 0.785
      & 0.870 & 0.861
      & 0.795 & 0.814
      & 0.795 & 0.817
      & 0.845 & \underline{0.847}
      & 0.677 & 0.669
      & 0.794 \\
    FAST-VQA~(TPAMI 2023)~~\cite{wu2022fast}
      & 0.765 & 0.793
      & 0.880 & 0.871
      & \textcolor{blue}{0.830} & 0.822
      & 0.869 & 0.870
      & 0.725 & 0.742
      & 0.631 & 0.677
      & 0.790 \\
    Dover~(ICCV 2023)~~\cite{wu2023exploring}
      & 0.797 & 0.821
      & 0.893 & 0.892
      & \textcolor{red}{0.835} & \textcolor{red}{0.857}
      & \textcolor{red}{0.885} & 0.879
      & 0.801 & 0.814
      & 0.647 & 0.728
      & 0.821 \\
    Modular-VQA~(CVPR 2024)~~\cite{wen2024modular}
      & \underline{0.810} & 0.834
      & \textcolor{red}{0.897} & \textcolor{blue}{0.895}
      & 0.803 & 0.839
      & 0.876 & \textcolor{red}{0.887}
      & 0.786 & 0.803
      & 0.696 & 0.703
      & 0.819 \\
    KVQ~(CVPR 2025)~~\cite{qu2025kvq}
      & \textcolor{blue}{0.814} & \textcolor{red}{0.846}
      & \textcolor{blue}{0.896} & \textcolor{red}{0.897}
      & \underline{0.820} & \underline{0.843}
      & 0.829 & 0.818
      & \textcolor{red}{0.890} & \textcolor{red}{0.892}
      & \underline{0.725} & \underline{0.732}
      & \textcolor{blue}{0.833} \\
   
    Q-ALIGN-VQA~(7B)~~\cite{wu2024q1}
      & 0.758 & 0.833
      & 0.883 & 0.882
      & 0.777 & 0.813
      & 0.865 & 0.876
      & 0.811 & 0.830
      & 0.608 & 0.675
      & 0.801 \\
    Q-ALIGN-Onealign~(7B)
      & 0.803 & \underline{0.836}
      & 0.888 & 0.885
      & \underline{0.773} & 0.829
      & \underline{0.876} & 0.878
      & 0.831 & \underline{0.847}
      & 0.611 & 0.681
      & 0.812 \\
       \cdashline{1-14}
    \textbf{\textsc{LLaVA-Assessor-GIGA}}
      & \textcolor{red}{0.817} & \textcolor{blue}{0.842}
      & 0.885 & \underline{0.893}
      & 0.808 & \textcolor{blue}{0.851}
      & \textcolor{blue}{0.878} & \textcolor{blue}{0.881}
      & \textcolor{blue}{0.858} & \textcolor{blue}{0.871}
      & \textcolor{red}{0.750} & \textcolor{red}{0.785}
      & \textcolor{red}{0.841} \\
    \hline
    \end{tabular}%
    }
    \vspace{-6pt}
    \label{tab:rating}
\end{table*}

\begin{table*}[tbp]
\vspace{-1em}
\centering
\small
\renewcommand\arraystretch{1.05}
\renewcommand\tabcolsep{12pt}
\caption{Performance on image quality scoring task. Per column: highest in \textcolor{red}{red}, second in \textcolor{blue}{blue}, third \underline{underlined}. [{$\spadesuit$}: Intra-dataset testing, $\diamondsuit$: Inter-dataset testing, \textit{Textit}: Out of domain test datasets]}
\vspace{-8pt}
\resizebox{\linewidth}{!}{%
\begin{tabular}{l|ccccccccccc}
\hline
\textbf{Datasets}
  & \multicolumn{2}{c}{\textbf{KonIQ}~\cite{hosu2020koniq}\textsuperscript{$\spadesuit$}}
  & \multicolumn{2}{c}{\textbf{SPAQ}~\cite{fang2020perceptual}\textsuperscript{$\spadesuit$}}
  & \multicolumn{2}{c}{\textbf{LIVE-C}~\cite{livechallenge}\textsuperscript{$\diamondsuit$}}
  & \multicolumn{2}{c}{\textbf{\textit{AGIQA-3K}~\cite{agiqa3k}\textsuperscript{$\diamondsuit$}}}
  & \multicolumn{2}{c}{\textbf{\textit{KADID-10K}}~\cite{lin2019kadid}\textsuperscript{$\diamondsuit$}}
  & \multirow{2}{*}{\textbf{Avg.}} \\ 
\cline{1-11}
  \textbf{Models}& SRCC & PLCC & SRCC & PLCC & SRCC & PLCC & SRCC & PLCC & SRCC & PLCC &  \\
\hline
NIMA~(TIP 2018)~~\cite{nima} 
  & 0.859 & 0.896 
  & 0.856 & 0.838 
  & 0.771 & 0.814 
  & 0.654 & 0.715 
  & 0.535 & 0.532 
  & 0.747 \\

DBCNN~(TCSVT 2020)~~\cite{dbcnn} 
  & 0.875 & 0.884 
  & 0.806 & 0.812 
  & 0.755 & 0.773 
  & 0.641 & 0.730 
  & 0.484 & 0.497 
  & 0.726 \\

HyperIQA~(CVPR 2020)~~\cite{hyperiqa} 
  & 0.906 & 0.917 
  & 0.788 & 0.791 
  & 0.749 & 0.772 
  & 0.640 & 0.702 
  & 0.468 & 0.506 
  & 0.724 \\

MUSIQ~(ICCV 2021)~~\cite{musiq} 
  & \underline{0.929} & \underline{0.924} 
  & 0.863 & 0.868 
  & 0.830 & 0.789 
  & 0.630 & 0.722 
  & 0.556 & 0.575 
  & 0.769 \\

CLIP-IQA+~(AAAI 2023)~~\cite{wang2023exploring} 
  & 0.895 & 0.909 
  & 0.864 & 0.866 
  & 0.805 & 0.832 
  & 0.685 & 0.736 
  & 0.654 & 0.653 
  & 0.790 \\

LIQE~(CVPR 2023)~~\cite{liqe} 
  & {0.928} & 0.912 
  & 0.833 & 0.846 
  & \textcolor{blue}{0.870} & 0.830 
  & 0.708 & \textcolor{blue}{0.772} 
  & 0.662 & 0.667 
  & 0.803 \\

Q-ALIGN-IQA~(ICML 2024)~~\cite{wu2024q1} 
  & \textcolor{blue}{0.940} & \textcolor{blue}{0.941} 
  & \underline{0.887} & \underline{0.886} 
  & \underline{0.860} & \textcolor{blue}{0.853} 
  & \underline{0.735} & \textcolor{blue}{0.772} 
  & \underline{0.684} & \underline{0.674} 
  & \underline{0.823} \\

DeQA~(CVPR 2025)~~\cite{you2025teaching} 
  & \textcolor{red}{0.943} & \textcolor{red}{0.953} 
  & \textcolor{blue}{0.933} & \textcolor{red}{0.936} 
  & 0.719 & \underline{0.852} 
  & \textcolor{red}{0.863} & 0.738 
  & \textcolor{red}{0.810} & \textcolor{blue}{0.719} 
  & \textcolor{red}{0.847} \\

\hdashline
\textbf{\textsc{LLaVA-Assessor-GIGA}} 
  & 0.887 & 0.892 
  & \textcolor{red}{0.934} & \textcolor{blue}{0.900} 
  & \textcolor{red}{0.874} & \textcolor{red}{0.864} 
  & \textcolor{blue}{0.759} & \textcolor{red}{0.810} 
  & \textcolor{blue}{0.764} & \textcolor{red}{0.727} 
  & \textcolor{blue}{0.837} \\
\hline
\end{tabular}}
\vspace{-10pt}
\label{tab:rating_image}
\end{table*}

\subsubsection{Training details}
Both \textit{LLaVA-Assessor-SFT} and \textit{LLaVA-Assessor-GIGA} are trained for \textbf{only one} complete epoch. The model training is conducted on $8$ A800-80G GPUs.
The training hyperparameters are detailed in Tab.~\ref{tab:hyperparam}. During training, the CE loss drops quickly and then stabilizes~(reaches approximately $0.5$ after $1,000$ steps)~(shown in Fig.~\ref{fig:trainingloss}).
\subsubsection{Inference details}
\label{scoring}
Following Q-ALIGN~\cite{wu2024q1}, we adopt the following mapping strategy to obtain the score from vocab logits  during quality scoring evaluation:
\begin{equation}
\mathcal{Q} = \sum_{i=1}^5 \omega_i \frac{e^{\mathcal{P}_{\textit{quality\_levels}[i]}}}{\sum_{i=1}^5 e^{\mathcal{P}_{\textit{quality\_levels}[i]}}},
\end{equation}
where \textit{quality\_levels} refers to the list of discrete quality levels: \textit{[High, Good, Fair, Poor, Low]}, and \(\mathcal{P}\) denotes the model's \textbf{output logit} of each respective quality level. These logits are subsequently normalized using the softmax function. The values \(\omega\) represent the weight factors assigned to the normalized probabilities of each quality level, given by $[0.9,0.7, 0.5, 0.3, 0.1]$. The resulting weighted sum of these probabilities produces the predicted quality score \(\mathcal{Q}\).
We directly call \textit{model.forward()} when making quality scoring.
The final output corresponds to the token after the last vision token, which represents the quality level.
For the quality interpretation task, we use \textit{model.generate()} with \textit{greedy search}~(setting \textit{do-sample} to \textit{False}) to ensure reproducibility.

\vspace{-15pt}
\subsection{Video Quality Scoring Performance}
We compare our model with several high-performing DNN-based~\cite{sun2022deep,li2022blindly,wu2022fast,wu2023exploring,wen2024modular,qu2025kvq}
and LMM-based~\cite{wu2024q1} quality scoring models on $6$ commonly used datasets including diverse scenarios. Apart from the pretrained \textit{Q-ALIGN}, and our model~(using mix training), all the comparing models are trained on the LSVQ~(train)~\cite{ying2021patch}. The evaluation metrics are the commonly used \textit{Pearson Linear Correlation Coefficient~(PLCC)} and  \textit{Spearman Rank Correlation Coefficient~(SRCC)}. 
The performance of the models on all datasets is presented in 
Tab.~\ref{tab:rating}. The experimental results demonstrate that the \textit{LLaVA-Assessor-GIGA} achieves \textit{Top-3} performance across almost all metrics of the $6$ test datasets. This demonstrates its superior performance in video quality scoring tasks.
\subsection{Image Quality Scoring Performance}
 We choose KonIQ-10K~\cite{hosu2020koniq}~(\textit{test set}), SPAQ~\cite{fang2020perceptual}~(\textit{test set}), LIVEC~\cite{livechallenge}, AGIQA-3K~\cite{agiqa3k}, and KADID-10K~\cite{lin2019kadid}~(\textit{test set}) as the test datasets. The experiment results are shown in Tab.~\ref{tab:rating_image}. In this task, despite a small performance gap compared to the latest proprietary IQA models on some datasets, our model still demonstrates outstanding performance.
\vspace{-10pt}
\subsection{Video Quality Interpreting Performance}
As the core evaluation task, we conduct comprehensive evaluation experiments on general video quality interpretation tasks. The experiment is performed on the \textit{Q-bench-video-test~(single)}~\cite{zhang2024q1}, consisting of $992$ unique question/video samples. It primarily focuses on evaluating the model's ability to describe general video quality, with questions that target quality issues perceptible to the average viewer. Additionally, it includes a small number of expert-level fine-grained video quality interpretation questions. The bench includes various question types: binary~(\textit{Binary}), multiple-choice~(single-answer)~(\textit{Multi.}), and open-ended questions~(\textit{Open}). These questions also cover diverse quality concerns, such as technical quality~(\textit{Tech.}), aesthetic quality~(\textit{Aes.}), temporal quality~(\textit{Temp.}), and AIGC video quality~(\textit{AIGC}).

Here, we conduct a comprehensive LMM performance evaluation, selecting the latest models~(including medium and large parameter sizes) from $3$ renowned open-source series~(\textit{QwenVL}~\cite{wang2024qwen2,bai2025qwen2}, \textit{InternVL}~\cite{chen2024expanding,internVL3,wang2024enhancing}, and \textit{LLaVA}~\cite{li2024llava}), along with other high-performing open-source models~\cite{ye2024mplug1} for comparison. For proprietary models, we choose the representative ones, including \textit{GPT-4o}~\cite{achiam2023gpt}, \textit{Gemini-2.5-Pro}~\cite{gemini2025gemini25}, \textit{Grok2-Vision}~\cite{xai2024grok2}, and \textit{Claude-3.7-Sonnet}~\cite{anthropic2025claude}. These experiments serve both as a comparison with our model and as a supplementary LMM evaluation on the corresponding benchmarks. The experimental results are summarized in Tab.~\ref{tab:overall}.

Results reveal that our model outperforms almost all other models with noticeable gain, particularly excelling in \textit{Tech., Aes., and Temp.}  aspects. This highlights our model's effectiveness in general video quality interpretation tasks.
\begin{table*}[t]\tiny
    \centering
    \renewcommand\arraystretch{1.05}
    \renewcommand\tabcolsep{9pt}
    \belowrulesep=0pt\aboverulesep=0pt

    \caption{Evaluation results on the \textit{Q-bench-video-test~(single)}. Per column: highest in \textcolor{red}{red}, second in \textcolor{blue}{blue}, third \underline{underlined}.}
 \vspace{-8pt}
    \resizebox{\linewidth}{!}{\begin{tabular}{l|ccc|cccc|c}
  
       \hline
     \textbf{Sub-categories} & \multicolumn{3}{c|}{\textbf{Question Types}} & \multicolumn{4}{c|}{\textbf{Quality Concerns}} &\multirow{2}{*}{\textit{Overall$\uparrow$}}
        \\ \cdashline{1-8}
        \multirow{1}{*}{\textbf{Models}} & \textit{Binary} $\uparrow$&\textit{Multi.} $\uparrow$& \textit{Open} $\uparrow$& \textit{Tech.} $\uparrow$&\textit{Aes.} $\uparrow$& \textit{Temp.}  $\uparrow$&\textit{AIGC} $\uparrow$& \\
    \cline{1-9}
     \multicolumn{9}{l}{\textit{{Performance of Open-sourced LMMs}}} \\ \hdashline
      mPLUG-Owl3-\textit{7B}~\cite{ye2024mplug1} &56.90\% & 57.14\% & 42.88\% & 53.40\% & 61.85\% & 50.34\% & 45.34\% & 52.06\% \\ 
      InternVL2.5-\textit{8B}~\cite{chen2024internvl25} &48.15\% & 39.37\% & 31.49\% & 39.06\% & 46.68\% & 42.52\% & 31.37\% & 39.50\% \\ 
       InternVL2.5-\textit{38B}&50.84\% & 46.69\% & 37.18\% & 44.57\% & 54.50\% & 48.13\% & 40.68\% & 44.72\%\\ 
        InternVL2.5-\textit{78B}&48.82\% & 46.34\% & 38.13\% & 43.52\% & 57.35\% & 47.28\% & 36.65\% & 44.28\%\\ 
        InternVL3-\textit{8B} &50.84\% & 44.25\% & 33.54\% & 41.82\% & 52.37\% & 45.07\% & 37.58\% & 42.67\%\\ 
         InternVL3-\textit{38B}&50.84\% & 46.34\% & 36.87\% & 44.49\% & 51.66\% & 45.24\% & 36.34\% & 44.50\%\\ 
          InternVL3-\textit{78B}&47.81\% & 50.17\% & 38.13\% & 45.79\% & 60.90\% & 46.43\% & 36.02\% & 45.17\%\\ 
      LLaVA-OneVision-\textit{7B}~\cite{li2024llava} &57.58\% & 48.78\% & 32.12\% & 44.98\% & 50.95\% & 45.07\% & 44.72\% & 45.83\% \\ 
       LLaVA-OneVision-\textit{72B}&52.19\% & 54.36\% & 34.34\% & 45.62\% & 54.74\% & 50.00\% & 46.58\% & 46.61\% \\ 
      
      Qwen2-VL-\textit{7B}~\cite{wang2024qwen2} &50.84\% & 55.75\% & 34.49\% & 46.03\% & 56.40\% & 50.17\% & 39.75\% & 46.67\%\\ 
      Qwen2-VL-\textit{72B} &61.62\% & \textcolor{blue}{66.90\%} & 39.24\% & \underline{55.19\%} & \underline{63.03\%} & 52.38\% & 50.93\% & 55.44\%\\ 
      
      Qwen2.5-VL-\textit{7B}~\cite{bai2025qwen2} &52.53\% & 49.48\% & 38.77\% & 46.68\% & 58.53\% & 46.09\% & 41.61\% & 46.72\%\\ 
       Qwen2.5-VL-\textit{72B} &62.73\% & 56.90\% & 41.31\% & 52.93\% & 62.49\% & 52.15\% & 46.79\% & 52.35\%\\ 
       \textbf{LLaVA-Assessor-SFT} &\underline{67.12\%} & 59.93\% & 39.56\% & \underline{55.19\%} & 56.87\% & \textcolor{blue}{57.99\%} & 43.79\% & \underline{55.56\%}\\ 
      \textbf{\textsc{LLaVA-Assessor-GIGA}}&\textcolor{red}{72.12\%} & \textcolor{red}{71.71\%} & 46.35\% & \textcolor{red}{61.60\%} & \textcolor{blue}{65.69\%} & \textcolor{red}{63.57\%} & \textcolor{blue}{68.65\%} & \textcolor{red}{62.94\%}\\ 
     
      \cline{1-9}
     \multicolumn{9}{l}{\textit{{Performance of Proprietary LMMs}}} \\ \hdashline
      GPT-4o~(24-11-20) &59.26\% & 51.57\% & \underline{47.63\%} & 51.38\% & 61.37\% & 51.87\% & 49.38\% & 52.72\%\\ 
      Gemini-2.5-Pro~(25-05-06) &\textcolor{blue}{69.70\%} & \underline{63.07\%} & \textcolor{red}{54.75\%} & \textcolor{blue}{61.43\%} & \textcolor{red}{68.48\%} & \underline{56.97\%} & \textcolor{red}{66.46\%} & \textcolor{blue}{62.33\%} \\ 
       Grok2-Vision~(24-11-23)        &55.22\% & 51.57\% & 41.93\% & 48.87\% & 54.98\% & 49.83\% & \underline{51.24\%} & 49.39\%\\ 
      Claude-3.7-Sonnet~(25-02-19)    &55.89\% & 42.51\% & \textcolor{blue}{48.89\%} & 49.03\% & 58.29\% & 46.09\% & 41.93\% & 49.17\%\\ 
         \hline
      \end{tabular}}
      \vspace{-10pt}

    \label{tab:overall}
\end{table*}
\begin{table*}\small
    \centering
    \renewcommand\arraystretch{1.0}
    \renewcommand\tabcolsep{14pt}
    \caption{Evaluation results on the \textit{LLVisionQA-test}. Per column: highest in \textcolor{red}{red}, second in \textcolor{blue}{blue}, third \underline{underlined}. [{$\spadesuit$}: Domain-specific LMM]}
    \vspace{-8pt}
    \resizebox{\linewidth}{!}{%
    \begin{tabular}{l|ccc|cc|cc|c}
    \hline
    \textbf{Sub-categories} & \multicolumn{3}{c|}{\textbf{Question Types}} & \multicolumn{4}{c|}{\textbf{Quality Concerns}} & \multirow{3}{*}{\textit{Overall $\uparrow$}} \\
    \cdashline{1-8}
    \multirow{2}{*}{\textbf{Models}} 
      & \multirow{2}{*}{\textit{Binary} $\uparrow$} 
      & \multirow{2}{*}{\textit{What} $\uparrow$} 
      & \multirow{2}{*}{\textit{How} $\uparrow$} 
      & \multirow{2}{*}{\textit{Technical} $\uparrow$} 
      & \multirow{2}{*}{\textit{Other} $\uparrow$}
      & \multicolumn{2}{c|}{\textit{In-context}}  \\
    &&&&&&\textit{Technical $\uparrow$} & \textit{Other $\uparrow$} \\
    \hline
    \multicolumn{9}{l}{\textit{Performance of Open-sourced LMMs}} \\ \hdashline
         mPLUG-Owl3-\textit{7B}
         & 78.72\% 
         & 79.77\% 
         & 67.45\% 
         & 73.44\% 
         & 71.74\% 
         & 71.19\% 
         & 84.89\% 
         & 74.21\% \\
         InternVL2.5-\textit{8B} 
         & 78.84\% 
         & 79.67\% 
         & 66.13\% 
         & 69.23\% 
         & 75.98\% 
         & 69.18\% 
         & 85.15\% 
         & 73.80\% \\
         InternVL2.5-\textit{38B} 
         & 79.37\% 
         & 82.64\% 
         & 68.93\% 
         & 73.70\% 
         & 78.52\% 
         & 72.26\% 
         & 86.31\% 
         & 76.98\% \\
         InternVL2.5-\textit{78B}
         & 79.92\% 
         & 80.47\% 
         & 70.57\% 
         & 73.32\% 
         & 77.80\% 
         & 75.00\% 
         & 85.55\% 
         & 77.05\% \\
         InternVL3-\textit{8B} 
         & 78.28\% 
         & 81.56\% 
         & 69.95\% 
         & 70.82\% 
         & 79.23\% 
         & 73.97\% 
         & \underline{86.69\%} 
         & 76.58\% \\
         InternVL3-\textit{38B} 
         & 79.92\% 
         & 83.29\% 
         & 70.57\% 
         & 74.47\% 
         & 78.99\% 
         & 75.00\% 
         & 86.31\% 
         & 77.92\% \\
         LLaVA-OneVision-\textit{7B} 
         & 79.12\% 
         & 78.19\% 
         & 69.73\% 
         & 70.06\% 
         & 76.54\% 
         & 73.11\% 
         & 83.01\% 
         & 74.68\% \\
         LLaVA-OneVision-\textit{72B}
         & 79.74\% 
         & \underline{85.24\%} 
         & 72.83\% 
         & 75.43\% 
         & \textcolor{blue}{82.10\%} 
         & 75.00\% 
         & 86.69\% 
         & 79.19\% \\
         Qwen2-VL-\textit{7B} 
         & 81.56\% 
         & 79.60\% 
         & 72.63\% 
         & 73.89\% 
         & 79.95\% 
         & 75.00\% 
         & 86.69\% 
         & 78.06\% \\
         Qwen2-VL-\textit{72B} 
         & \underline{81.93\%} 
         & \textcolor{blue}{85.24\%} 
         & \textcolor{blue}{75.30\%} 
         & \textcolor{red}{77.92\%} 
         & \textcolor{red}{82.10\%} 
         & 79.10\% 
         & 86.31\% 
         & \textcolor{blue}{80.67\%} \\
         Qwen2.5-VL-\textit{7B} 
         & 80.47\% 
         & 84.81\% 
         & 69.95\% 
         & 76.19\% 
         & 79.47\% 
         & 77.39\% 
         & 82.12\% 
         & 78.39\% \\
         Qwen2.5-VL-\textit{72B} 
         & 81.38\% 
         & 84.16\% 
         & \textcolor{red}{75.92\%} 
         & 76.96\% 
         & \underline{80.90\%} 
         & \textcolor{blue}{80.13\%} 
         & \textcolor{blue}{87.07\%} 
         & \underline{80.46\%} \\
         Seed-Bagel-\textit{7B}~\cite{deng2025emerging} 
         & 79.92\% 
         & \textcolor{red}{85.68\%} 
         & \underline{74.89\%} 
         & \textcolor{blue}{77.92\%} 
         & 78.04\% 
         & 78.76\% 
         & \textcolor{red}{88.97\%} 
         & 80.06\% \\
         Janus-Pro-\textit{7B}~\cite{chen2025janus} 
         & 64.23\% 
         & 68.32\% 
         & 59.46\% 
         & 55.27\% 
         & 68.73\% 
         & 57.87\% 
         & 80.22\% 
         & 63.94\% \\
          Q-Instruct-\textit{7B}~\cite{wu2024q} \textsuperscript{$\spadesuit$}
         & 64.23\% 
         & 68.32\% 
         & 59.46\% 
         & 55.27\% 
         & 68.73\% 
         & 57.87\% 
         & 80.22\% 
         & 63.94\% \\
          \textbf{\textsc{LLaVA-Assessor-GIGA}} 
         & \textcolor{red}{83.94\%} 
         & 83.29\% 
         & 68.72\% 
         & \underline{77.35\%} 
         & 77.89\% 
         & \textcolor{red}{82.87\%} 
         & 83.74\% 
         & \textcolor{red}{80.79\%} \\
    \hline
    \multicolumn{9}{l}{\textit{Performance of Proprietary LMMs}} \\ \hdashline
        GPT-4o
         & \textcolor{blue}{82.48\%} 
         & 83.94\% 
         & 70.16\% 
         & 76.00\% 
         & 80.19\% 
         & \underline{79.45\%} 
         & 82.12\% 
         & 78.92\% \\
          Gemini-2.5-Pro &
        81.56\% & 87.41\% & 68.31\% & 75.23\% & 78.99\% & 84.93\% & 80.22\% & 79.06\%\\
         
         Grok2-vision
         & 74.45\% 
         & 78.74\% 
         & 67.28\% 
         & 70.63\% 
         & 75.65\% 
         & 69.17\% 
         & 80.22\% 
         & 73.44\% \\
         Claude-3.7-Sonnet
         & 74.08\% 
         & 78.95\% 
         & 66.46\% 
         & 70.05\% 
         & 75.65\% 
         & 68.83\% 
         & 79.84\% 
         & 73.11\% \\

    \hline
    \end{tabular}}
    \vspace{-10pt}
    \label{tab:perception}
\end{table*}
\begin{table*}\small
    \centering
    \renewcommand\arraystretch{1.0}
    \renewcommand\tabcolsep{16pt}
    \caption{Results on \textit{LLVisionQA-pair-test}. Per column: highest in \textcolor{red}{red}, second in \textcolor{blue}{blue}, third \underline{underlined}.}
    \vspace{-8pt}
    \resizebox{\linewidth}{!}{\begin{tabular}{l|ccc|cc|cc|c}
    \hline
        \textbf{Sub-categories} 
        & \multicolumn{3}{c|}{\textbf{Question Types}} 
        & \multicolumn{2}{c|}{\textbf{Low-level Concerns}} 
        & \multicolumn{2}{c|}{\textbf{Pairwise Concerns}} 
        & \multirow{2}{*}{\textit{Overall $\uparrow$}} \\ 
    \cdashline{1-8}
        \textbf{Models}  
        & \textit{Binary $\uparrow$}
        & \textit{What $\uparrow$} 
        & \textit{How $\uparrow$} 
        & \textit{Distortion $\uparrow$} 
        & \textit{Other $\uparrow$} 
        & \textit{Compare $\uparrow$}  
        & \textit{Joint $\uparrow$}  \\ 
    \hline
    \multicolumn{9}{l}{\textit{Performance of Open-sourced LMMs}} \\ \hdashline
    mPLUG-Owl3-\textit{7B} 
      & 54.21\% 
      & 43.38\% 
      & 45.32\% 
      & 49.57\% 
      & 45.67\% 
      & 48.32\% 
      & 48.88\% 
      & 48.44\% \\

    InternVL2.5-\textit{8B}
      & 51.94\% 
      & 29.78\% 
      & 53.84\%
      & 42.01\% 
      & 55.71\% 
      & 46.26\% 
      & 49.09\% 
      & 47.08\% \\

    InternVL3-\textit{8B}
      & 70.36\% 
      & 28.13\% 
      & 35.98\% 
      & 44.08\% 
      & 57.43\% 
      & 47.02\% 
      & 51.11\% 
      & 47.94\% \\

    LLaVA-OneVision-\textit{7B}
      & 60.09\% 
      & 45.42\% 
      & 50.86\%
      & 53.09\% 
      & 58.82\% 
      & 54.52\%
      & 55.55\% 
      &52.75\% \\

    Qwen2-VL-\textit{7B}
      & 60.24\% 
      & 47.46\% 
      & 48.78\% 
      & 52.81\%  
      & 53.97\% 
      & 51.42\% 
      & 59.11\% 
      & 53.15\% \\

    Qwen2.5-VL-\textit{7B}
      & 58.07\% 
      & 36.61\% 
      & 48.44\% 
      & 47.74\% 
      & 51.90\% 
      & 45.73\% 
      & 60.00\% 
      & 48.94\% \\

    \textbf{\textsc{LLaVA-Assessor-GIGA}}
      & \textcolor{blue}{80.43\%} 
      & 64.02\% 
      & \underline{72.78\%}
      & \textcolor{blue}{79.88\%}
      & \textcolor{red}{62.03\%} 
      & \underline{72.24\%}
      & 85.89\% 
      & \textcolor{blue}{74.33\%}\\

    \hline
    \multicolumn{9}{l}{\textit{Performance of Proprietary LMMs}} \\ \hdashline
    GPT-4o
      & \textcolor{red}{82.07\%} 
      & \textcolor{red}{72.60\%} 
      & \textcolor{red}{74.12\%} 
      & \textcolor{red}{80.90\%} 
      & \textcolor{blue}{61.78\%} 
      & \textcolor{red}{72.83\%} 
      & 87.71\% 
      & \textcolor{red}{76.82\%} \\

    Gemini-2.5-Pro  
      & 76.67\% 
      & 60.28\% 
      & \textcolor{blue}{73.25\%} 
      &  \underline{75.32\%} 
      & \underline{60.93\%} 
      & \textcolor{blue}{72.65\%} 
      & \underline{88.86\%} 
      & 71.46\% \\
     Grok2-Vision
      & 78.28\% 
      & \textcolor{blue}{65.06\%} 
      & 70.28\% 
      & 73.33\%
      & 59.34\%
      & 70.94\%
      & \textcolor{red}{89.47\%} 
      & 71.92\% \\

    Claude-3.7-Sonnet
      & \underline{78.53\%} 
      & \underline{64.04\%} 
      &  71.24\% 
      &73.69\% 
      & 57.72\% 
      & 71.32\%
      & \textcolor{red}{89.47\%} 
      & \underline{72.02\%} \\

    \hline
     \end{tabular}}
    \vspace{-18pt}
    \label{tab:perception_pair}
\end{table*}

\begin{table}\small
    \centering
    \renewcommand\arraystretch{1.05}
    \renewcommand\tabcolsep{11pt}
    \belowrulesep=0pt\aboverulesep=0pt
    \caption{Results on the \textit{Q-bench-video-test~(pair)}. Per column: highest in \textcolor{red}{red}, second in \textcolor{blue}{blue}, third \underline{underlined}.}
    \vspace{-8pt}
    \resizebox{\linewidth}{!}{\begin{tabular}{l|cccc}
    \toprule
        \textbf{Sub-categories} & \multicolumn{4}{c}{\textbf{Video Pairs}}  \\ \hdashline
        \multirow{2}{*}{\textbf{Models}} & \multirow{2}{*}{\textit{Joint $\uparrow$}} & \textit{Compare} & \textit{Compare}  & \multirow{2}{*}{\textit{Overall $\uparrow$}} \\
        &&\textit{-fine $\uparrow$}&\textit{-coarse $\uparrow$}&  \\ \hline
        \multicolumn{5}{l}{\textit{Performance of Open-sourced LMMs}} \\ \hdashline
        InternVL2.5-\textit{8B}  &48.85\% &51.10\% &49.20\% &49.79\%\\
        InternVL3-\textit{8B}&55.73\% & 40.40\% & 51.84\% & 49.72\% \\
        LLaVA-OneVision-\textit{7B} & 51.56\% & 48.43\% & 49.89\% & 53.48\% \\
        Qwen2.5-VL-\textit{7B} &59.05\% & 58.81\% & 54.72\% & 55.65\%\\
         \textbf{\textsc{LLaVA-Assessor-GIGA}}& \textcolor{red}{73.00\%} & \textcolor{blue}{65.81\%} & 61.72\% & \textcolor{blue}{66.74\%}\\
        \hline
        \multicolumn{5}{l}{\textit{Performance of Proprietary LMMs}} \\ \hdashline
        GPT-4o  & \underline{57.00\%} & \underline{62.97\%} & \textcolor{blue}{73.44\%} & 66.29\%\\
         Gemini-2.5-Pro & \textcolor{blue}{61.00\%} & \textcolor{red}{66.63\%} & \underline{69.28\%} & \textcolor{red}{68.88\%} \\
         
         Grok2-Vision &48.00\% & 60.00\% & 63.84\% & 59.16\%\\
         Claude-3.7-Sonnet &53.00\% & 60.59\% & \textcolor{red}{77.28\%} & \underline{66.29\%} \\
         
       \bottomrule
    \end{tabular}}
    \vspace{-15pt}
    \label{tab:pair_test}
\end{table}
\begin{table}[t]\small
    \centering
    \renewcommand\arraystretch{1.1}
    \renewcommand\tabcolsep{1.2pt}
    \caption{Ablation study results. \textbf{PD} denotes the prompt disentanglement strategy. \textbf{\textsc{Scoring}} performance is reported as the average of SRCC and PLCC on corresponding task datasets, while \textbf{\textsc{Interpreting}} performance is evaluated on the \textit{LLVisionQA-test} and \textit{Q-bench-video-test~(single)} for image and video, respectively. Per column: highest in \textcolor{red}{red}.}
    \vspace{-8pt}
    \resizebox{\linewidth}{!}{\begin{tabular}{c|cccc|cc|cc}
   
    \hline
      \textbf{\textsc{LLaVA-}}  &  \textbf{Image} & \textbf{Image} & \textbf{Video} & \textbf{Video}& \multicolumn{2}{c|}{\textbf{\textsc{Scoring}}} & \multicolumn{2}{c}{\textbf{Interpreting}}\\
      
      \textbf{\textsc{Assessor}}  &  \textbf{Scoring} & \textbf{Interpreting} & \textbf{Scoring} & \textbf{Interpreting}& \textbf{Image} &\textbf{Video}& \textbf{Image} &\textbf{Video}\\
    \hdashline
        - & $\times$ & \checkmark &$\times$& \checkmark & 0.735 &0.629&78.28\%&62.89\%\\
        - &\checkmark & $\times$&\checkmark  & $\times$&0.851&0.818&/&/\\
        - & $\times$ & $\times$ & \checkmark &\checkmark& 0.778 &0.825 &69.62\%&59.25\%\\
         -  & \checkmark &\checkmark& $\times$ & $\times$& 0.853 &0.653 &74.56\%&43.76\%\\
        - & $\times$ & $\times$ & $\times$ &\checkmark& 0.632&0.758 &60.12\%&59.11\%\\
        - &$\times$ &\checkmark& $\times$ &$\times$& 0.713 &0.511 &75.32\%&45.89\%\\
        \hdashline
        \textit{w/o} PD & \checkmark & \checkmark & \checkmark & \checkmark& \textcolor{red}{0.841} &0.838 &72.43\%&58.23\% \\
        \textit{w} PD & \checkmark & \checkmark & \checkmark & \checkmark& 0.837  &\textcolor{red}{0.841}&\textcolor{red}{80.79\%}&\textcolor{red}{62.94\%}\\
    \hline   
    \end{tabular}}
    \label{tab:ablation}
    \vspace{-12pt}
\end{table}
\begin{figure}[h]
    \centering
    \includegraphics[width=\linewidth]{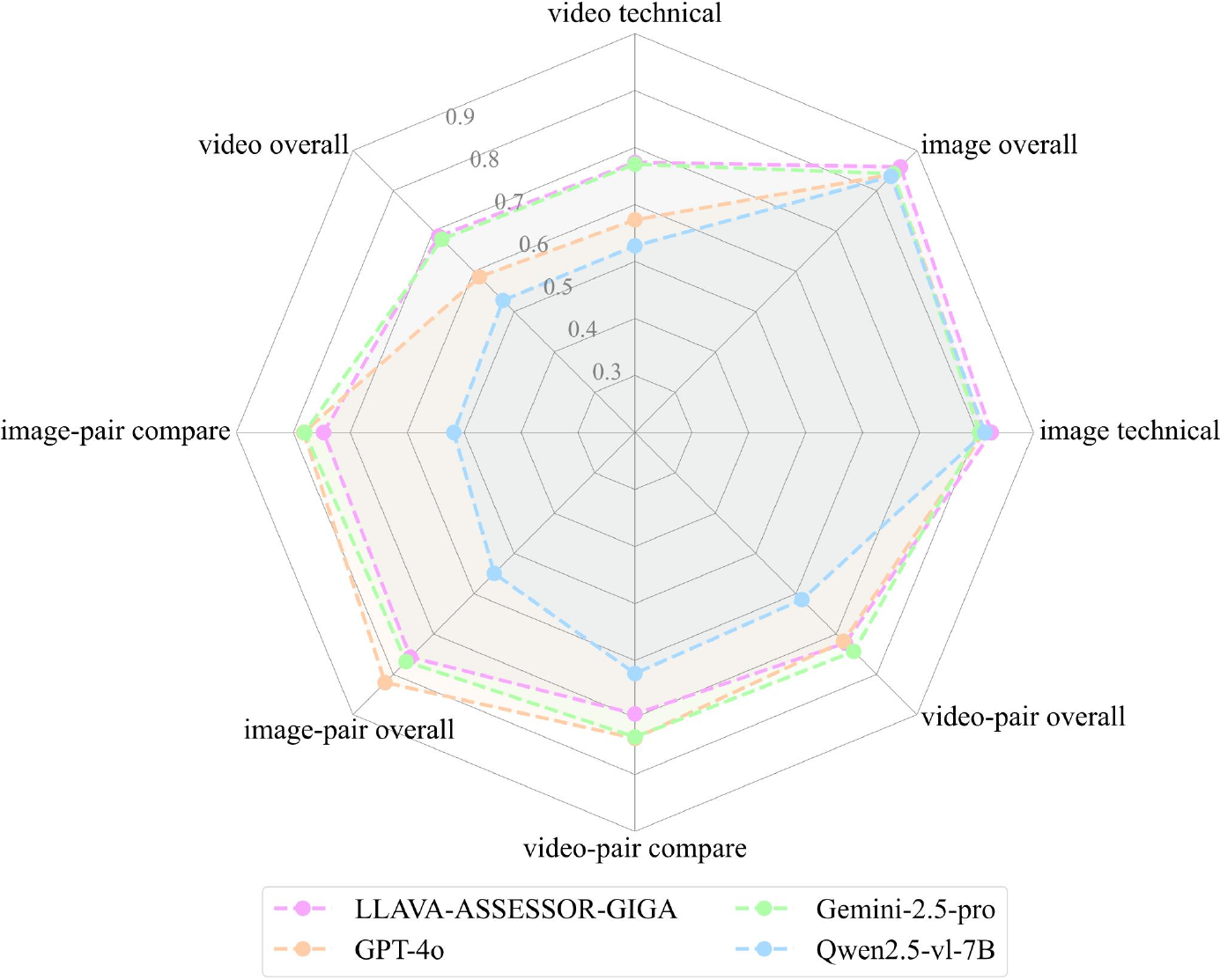}
    \caption{The overview visualization of the performance across a series of LMMs and our \textsc{LLaVA-Assessor-GIGA}.}
    \label{fig:performance}
    \vspace{-0.2cm}
\end{figure}
\subsection{Image Quality Interpreting Performance}
 We choose the commonly used \textit{LLVisionQA-test}~\cite{wu2024qbench}~($1,495$ questions) for image quality interpreting evaluation. Similar to \textit{Q-bench-video}, this benchmark also includes a variety of question types and tasks focused on analyzing technical~(distortion), aesthetic, AIGC, and in-context quality.
In addition to the comparison LMMs used in the video quality interpretation experiments, we also select two unified image generation and understanding large models~\cite{deng2025emerging,chen2025janus} for comparison.

The performance of LMMs on the \textit{LLVisionQA-test} is shown in Tab.\ref{tab:perception}, from which we have some observations: The overall performance of the \textit{Llava-Assessor-GIGA} surpasses that of open-sourced LMMs of the same size. Although it does not outperform some high-performing proprietary models or larger-scale LMMs, the performance gap is relatively marginal. Specifically, \textit{Llava-Assessor} shows a clear advantage in multiple-choice questions and technical dimensions, which aligns with the main composition of the training data.

\vspace{-8pt}
\subsection{Comparing and Joint Analysis Performance}
For low-level visual quality assessment LMMs, the ability to compare and jointly analyze multiple visual stimuli is also a compelling functionality. Since our model is trained exclusively on data containing a single visual signal, it is crucial to evaluate whether \textbf{large-scale training on this can stimulate the model's ability to perceive and interpret multiple visual signals}. To investigate this, we select the \textit{LLVisionQA-pair-test}~\cite{zhang2024qbench}~($1,000$ questions) and \textit{Q-bench-video-pair}~\cite{zhang2024q1}~($286$ questions) for evaluation on images and videos, respectively.

The results demonstrate that our model significantly outperforms open-source LMMs of the same size in the image/video visual quality compare/joint analysis tasks, achieving comparable performance to proprietary models. Here, considering the substantial cost associated with deploying these proprietary models, our model also provides a \textit{highly cost-effective alternative}. This highlights the impact of large-scale, diverse data post-training on the model’s ability to make the few-shot transfer from single visual signals to multiple visual signals.

Finally, Fig.~\ref{fig:performance} provides an intuitive visualization of the performance differences between our model and several representative LMMs across the above-mentioned tasks.

\vspace{-10pt}
\subsection{Ablation Study}
\label{ablation}
We conduct ablation experiments to evaluate the contribution of each training data component to the final model performance. Specifically, for each ablated variant, we remove the corresponding data subset and retrain the model using the same training strategy. The experimental results are reported in Tab.~\ref{tab:ablation}. We summarize the key observations as follows. \textbf{1)} Quality interpretation training directly improves the performance of quantitative quality scoring tasks. In particular, the results in the $1st$, $5th$, and $6th$ rows show that training with the interpretation task alone can still achieve acceptable performance on scoring tasks. \textbf{2)} The comparison between the $2nd$ and $7th$ rows, as well as that between the $1st$ and $7th$ rows, demonstrates that mixed training with both scoring and interpretation tasks can jointly enhance the performance of both tasks, with a more pronounced improvement in quality scoring. \textbf{3)} The results in the $3rd$ and $4th$ rows, compared with those in the $7th$ row, indicate that joint training with both image and video data contributes to performance improvements across different modalities. \textbf{4)} Large-scale SFT on a single scoring task or a single modality may cause the model to overfit to the corresponding task or modality, leading to degradation or even collapse of its capability on other tasks or modalities. The single-task and single-modality training results, as shown in the $2nd$, $3rd$, and $4th$ rows, restrict the model's generalization ability across modalities and tasks, while also weakening its capacity to follow complex instructions. Furthermore, based on the results reported in rows $3$–$6$, we observe that, \textbf{under our training settings, the generalization from image to video is stronger than that from video to image}.

Subsequently, we conduct experiments to verify \textbf{the effect of prompt disentanglement}~(denoted as \textbf{PD}). We add the prompts with complete semantic guidance to the quality scoring training data for the version without prompt disentanglement. The results of the experiment are also presented in the lower part of Tab.~\ref{tab:ablation}. It is evident that when using the prompt disentanglement setting, the model shows a significant performance gain in both the image and video quality interpreting tasks.

\vspace{-10pt}
\subsection{Data Scaling Effects}
\begin{figure}
    \centering
    \includegraphics[width=\linewidth]{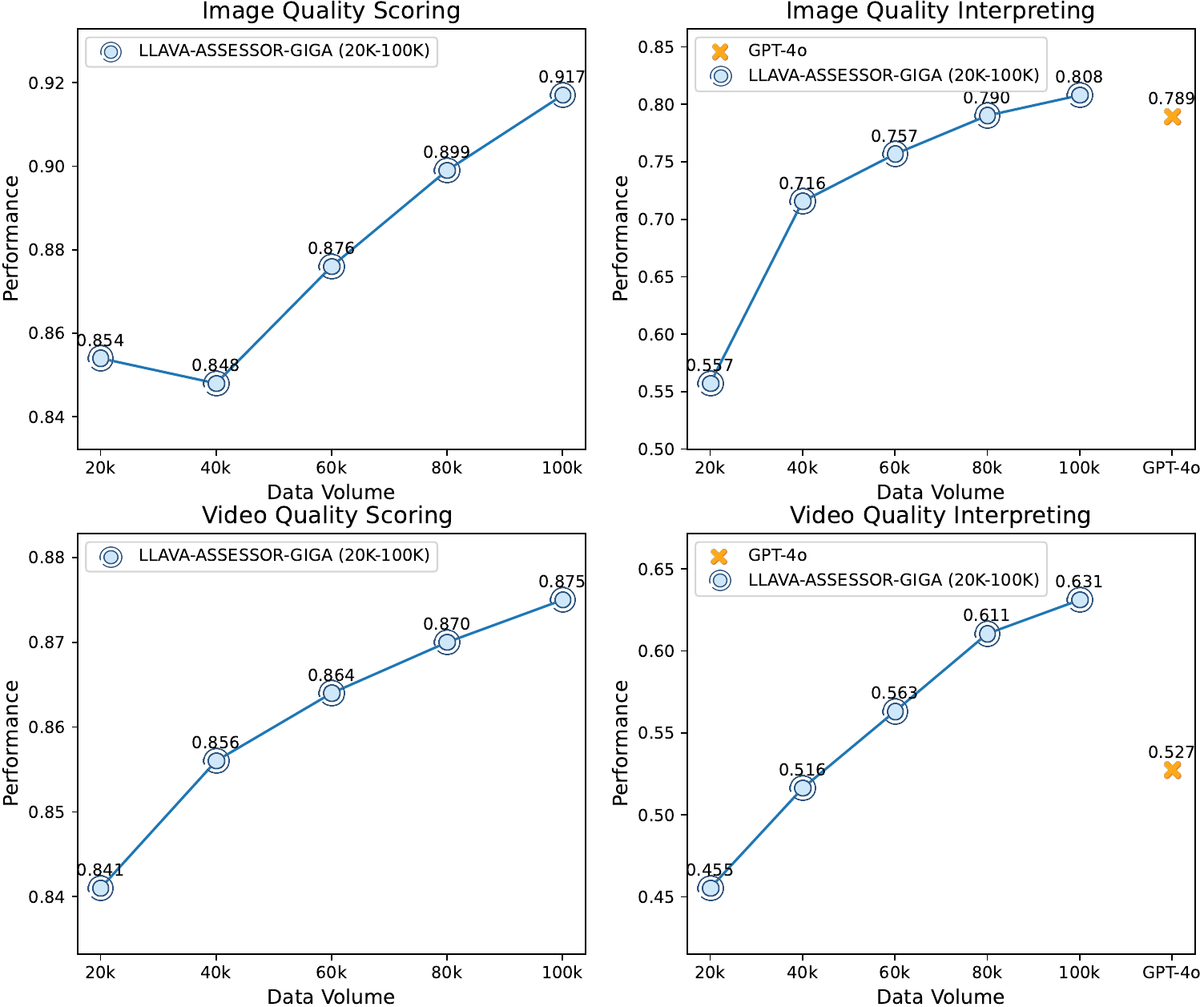}
    \vspace{-15pt}
    \caption{Visualization of the data-scaling effects, which shows the performance trends for the four tasks with the equal-proportional training data scaling-up. The image quality scoring task is evaluated on the \textit{SPAQ} test set, while the video quality scoring task is assessed on \textit{KoNViD-1K}. The performance for image and video quality interpreting is measured  on \textit{overall} performance on the \textit{LLVisionQA-test} and \textit{Q-bench-video-test~(single)}, respectively.}
    \label{fig:data_scaling}
    \vspace{-0.2cm}
\end{figure}

Additionally, we investigate the data-scaling effect on training by selecting $5$ data amount levels: $20K$, $40K$, $60K$, $80K$, and $100K$~(the full data scale). At each of these levels, the proportions of data for image quality scoring, image quality interpreting, video quality scoring, and video quality interpreting are kept consistent for fair comparison. Using the same training strategy for each level, the data-scaling effects for the model across the four tasks are presented in Fig.~\ref{fig:data_scaling}. 

We have several findings: \textbf{1}) The model performance for all four tasks improves progressively with the increase in training data, indicating that the marginal effect of the training dataset is relatively minimal. \textbf{2}) The performance of the image quality interpreting task reaches near saturation earlier compared to video quality interpreting, with the data scaling up. This could be due to $2$ factors: first, the base model~(\textit{LLaVA-OneVision}) may have already been trained on image quality interpreting data, thus already performing well in this task; second, the model's ability to perceive and interpret image visual quality is generally easier than for video, thus less amount of data is needed to achieve good performance. \textbf{3}) The performance of both image and video quality interpreting tasks becomes more marginal in the later stages of data scaling, suggesting that the model's parameter capacity may have been near saturation with respect to the data amount.

\vspace{-10pt}
\subsection{Computational Cost Analysis}

In this section, we focus on the computational cost
of \textit{Llava-Assessor-GIGA} during both training and inference stages.

\textbf{1) Training Cost}: During training, the VRAM usage on 8 A100 GPUs is around 560GB. The time consumption for a single epoch training is around $130h$.

\textbf{2) Inference Cost}: For an $8$s video, the input uses $1$fps and $384\times384$ resolution. The latency for the quality scoring is $0.89$s, with a VRAM usage of approximately $17$GB. For the quality interpretation, tasked with answering ``Please describe the video quality in detail", the latency is about $2$s, with a VRAM usage of approximately $20$GB. Overall, the inference cost is comparable to that of typical medium-sized LMMs.

\section{Conclusion}
In this paper, we introduce \textit{\textbf{LLaVA-Assessor}}, a \textbf{unified data curation and LMM training system} for multi-modal visual quality assessment. Starting from the classical HVS perception--decision framework, we map these two stages to two low-level machine vision tasks, namely quality interpretation and quality scoring, and define versatility'' and efficiency'' as the primary design objectives. Architecturally, we integrate \textit{SlowFast} to enable efficient video motion modeling and design a modality-aware gated structure that is selectively activated according to the input modality. For data construction and training, we propose a rigorous video-centric human annotation protocol, through which human experts annotate a large-scale MIDB for training \textit{LLaVA-Assessor-SFT}. We then employ \textit{LLaVA-Assessor-SFT} as the primary annotator and combine it with distortion synthesis, rejection sampling, LMM-as-a-judge, and human-in-the-loop scrutiny to construct video-centric machine-synthesized data. Furthermore, we introduce a simple yet effective prompt disentanglement strategy to alleviate target confusion between quality scoring and quality interpretation during mixed-task training. Based on this strategy, we perform multi-modal and multi-task joint training to obtain \textit{LLaVA-Assessor-GIGA}, which achieves superior performance on multiple benchmarks for both quality scoring and quality interpretation. Our results show that, with a rigorously structured annotation paradigm and a large-scale MIDB dominated by machine-synthesized data, foundation LMMs can achieve strong performance in low-level visual quality assessment. This study provides evidence for the feasibility of leveraging \textbf{machine vision} toward \textbf{automated visual quality assessment}.
\newpage
{
    \bibliographystyle{IEEEtran}
    \bibliography{ref}
}
\begin{IEEEbiography}[{\includegraphics[width=1.251in,height=1.25in,clip,keepaspectratio]{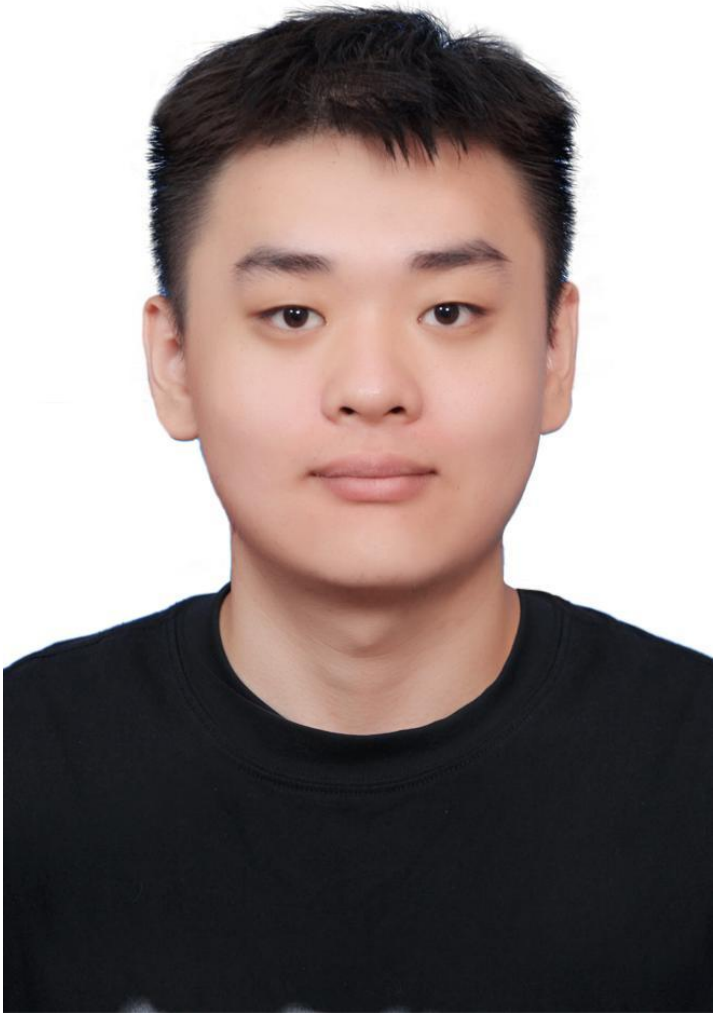}}]{Ziheng Jia}
received the B.E. degree from the Harbin Institute of Technology, Harbin, China, in 2023. He is currently working toward a Ph.D. degree with the Institute of Image Communication and Network Engineering, Shanghai Jiao Tong University, Shanghai, China. His current research interest is primarily in visual quality assessment.
\end{IEEEbiography}
\begin{IEEEbiography}[{\includegraphics[width=1.25in,height=1.25in,clip,keepaspectratio]{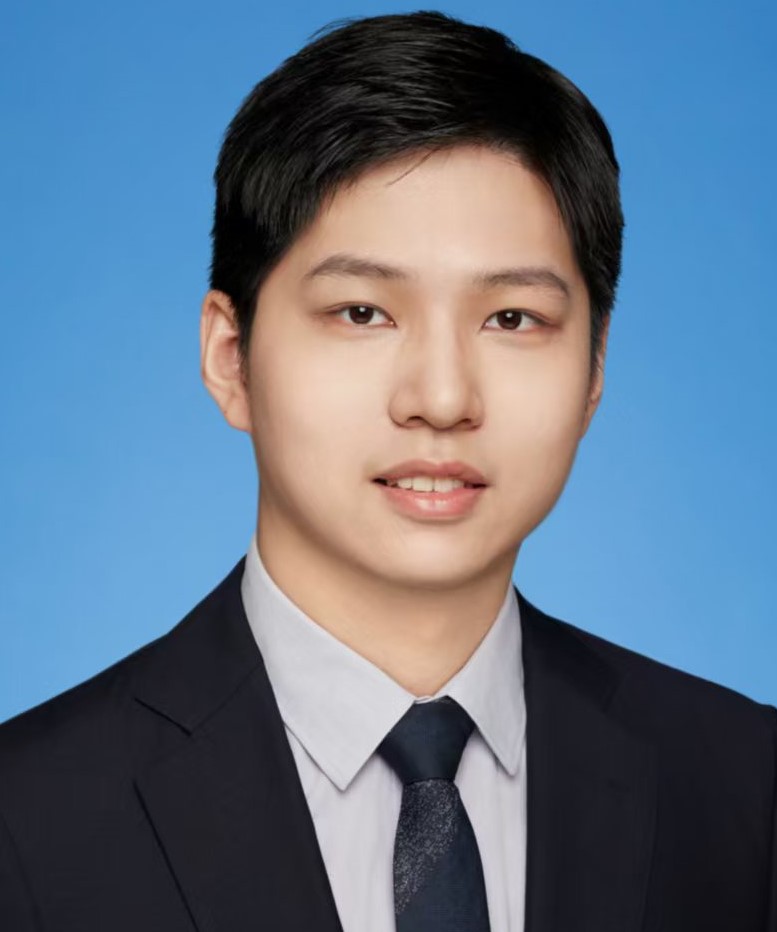}}]{Zicheng Zhang}
received his B.E. and Ph.D. degrees from Shanghai Jiao Tong University, Shanghai, China, in 2020 and 2025, respectively. He currently leads the Large Models Evaluation Group at the Center for AI Evaluation, Shanghai AI Laboratory. His work has been recognized with a Best Paper Nomination at ACM MM 2024 and the 2025 Scott Helt Memorial Award from the IEEE Broadcast Technology Society. He has served as an Area Chair or Editor for ICME, ACM MM, and Displays, and regularly reviews for leading journals, including IEEE TPAMI, TIP, TMM, and TCSVT. His research interests include visual quality assessment, low-level vision, and large multimodal models.
\end{IEEEbiography}
\begin{IEEEbiography}[{\includegraphics[width=1.25in,height=1.25in,clip,keepaspectratio]{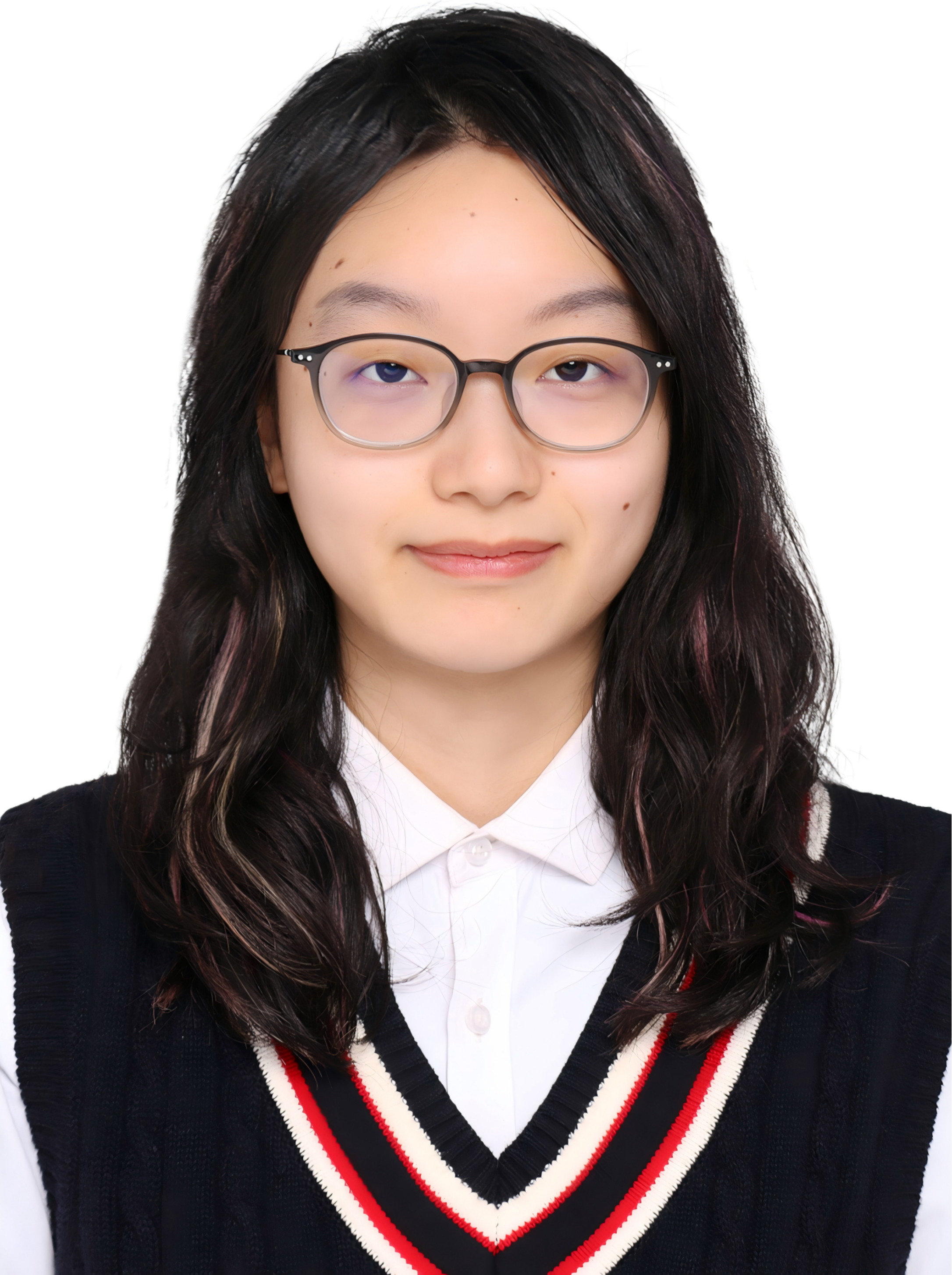}}]{Jiaying Qian}
is currently working toward the B.E. degree in Electronic Science and Technology at Shanghai Jiao Tong University, Shanghai, China. She will soon pursue the Ph.D. degree with the Institute of Image Communication and Network Engineering, Shanghai Jiao Tong University, Shanghai, China. Her research interests include image quality assessment.
\end{IEEEbiography}
\begin{IEEEbiography}[{\includegraphics[width=1.25in,height=1.25in,clip,keepaspectratio]{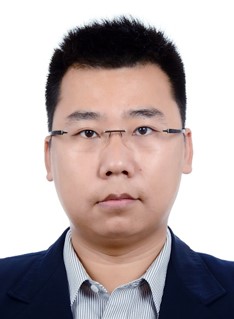}}]{Guangtao Zhai}
(Fellow 24, IEEE) received the B.E. and M.E. degrees from Shandong University, Shandong, China, in 2001 and 2004, respectively, and the Ph.D. degree from Shanghai Jiao Tong University, Shanghai, China, in 2009, where he is currently a Research Professor with the Institute of Image Communication and Information Processing. From 2008 to 2009, he was a Visiting Student with the Department of Electrical and Computer Engineering, McMaster University, Hamilton, ON, Canada, where he was a Post-Doctoral Fellow from 2010 to 2012. From 2012 to 2013, he was a Humboldt Research Fellow with the Institute of Multimedia Communication and Signal Processing, Friedrich Alexander University of Erlangen-Nuremberg, Germany. He received the Award of National Excellent Ph.D. Thesis from the Ministry of Education of China in 2012. His research interests include multimedia signal processing and perceptual signal processing.
\end{IEEEbiography}
\begin{IEEEbiography}[{\includegraphics[width=1.25in,height=1.25in,clip,keepaspectratio]{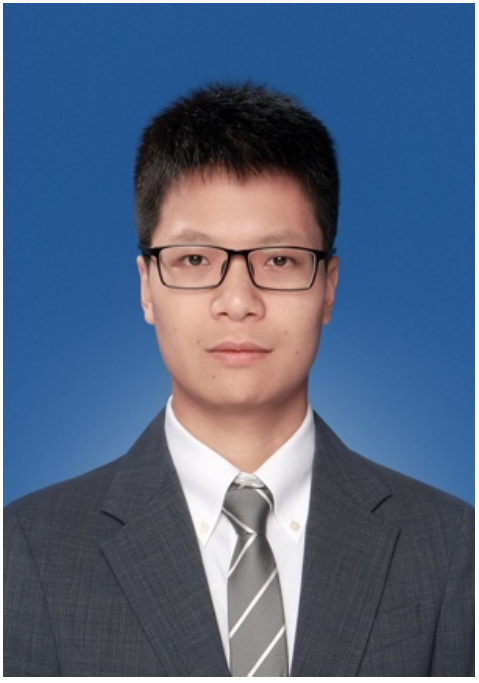}}]{Xiongkuo Min}
(Member, IEEE) received the B.E. degree from Wuhan University, Wuhan, China, in 2013, and the Ph.D. degree from Shanghai Jiao Tong University, Shanghai, China, in 2018, where he is currently a tenure-track Associate Professor with the Institute of Image Communication and Network Engineering. From Jun. 2018 to Sept. 2021, he was a Postdoc at Shanghai Jiao Tong University. From Jan. 2016 to Jan. 2017, he was a visiting student at the University of Waterloo. From Jan. 2019 to Jan. 2021, he was a visiting scholar at The University of Texas at Austin and the University of Macau. He received the Best Paper Runner-up Award of IEEE Transactions on Multimedia in 2021, the Best Student Paper Award of IEEE International Conference on Multimedia and Expo~(ICME) in 2016, the Best Paper Award of IEEE International Symposium on Broadband Multimedia Systems and Broadcasting~(BMSB) in 2022, and several first-place awards of grand challenges held at IEEE ICME and ICIP. His research interests include image/video/audio quality, quality of experience, multimedia, image/video processing, and computer vision.
\end{IEEEbiography}

\end{document}